\documentclass[10pt,twocolumn,letterpaper]{article}

\usepackage{cvpr}              

\usepackage{caption}
\usepackage{lipsum}
\usepackage[x11names, dvipsnames, table, svgnames]{xcolor} 
\usepackage{array}    
\usepackage{colortbl} 
\usepackage{amsfonts}
\usepackage{multirow}
\usepackage{tabularx}
\usepackage{array}
\usepackage{longtable}
\usepackage{booktabs}  
\usepackage{arydshln} 
\usepackage{subcaption}
\usepackage{amssymb}
\usepackage{float}
\usepackage{bbm}
\usepackage{tikz} 
\usepackage[accsupp]{axessibility}  

\definecolor{textgreen}{RGB}{0, 176, 80}
\definecolor{finegreen}{RGB}{0, 178, 14}
\definecolor{coarsegreen}{RGB}{195,250,150}
\definecolor{fineblue}{RGB}{90,169,230}
\definecolor{coarseblue}{RGB}{187,222,251}
\definecolor{fineorange}{RGB}{255,134,0}
\definecolor{suborange}{RGB}{0, 116, 217}
\definecolor{coarseorange}{RGB}{252, 222, 156}
\definecolor{darkblue}{rgb}{0,0,0.5}
\definecolor{darkred}{rgb}{0.8,0,0}
\definecolor{darkgreen}{rgb}{0,0.8,0}
\definecolor{refcolor}{HTML}{487AB8}

\definecolor{mybasic}{HTML}{caf0f8} 
\definecolor{mysub}{HTML}{90e0ef}  
\definecolor{myfine}{HTML}{0074D9} 

\usepackage{xspace}

\usepackage{lipsum}

\usepackage{titletoc}
\usepackage{minitoc}
\usepackage{etoc}
\etocsettocstyle{\section*{Contents}}{}

\definecolor{fig6red}{HTML}{FF4136}
\definecolor{fig6blue}{HTML}{0074D9}

\newcommand{\redcircle}{\tikz[baseline=-0.5ex]\draw[fill=fig6red, fill opacity=0.7,draw=black] (0,0) circle (0.5ex);}

\newcommand{\bluecircle}{\tikz[baseline=-0.5ex]\draw[fill=fig6blue, fill opacity=0.7,draw=black] (0,0) circle (0.5ex);}

\newcommand{\dataIN}{ImageNet\xspace}
\newcommand{\dataCUB}{CUB\xspace}
\newcommand{\dataAir}{Aircraft\xspace}
\newcommand{\dataInat}{iNat21-mini\xspace}

\newcommand{\dataINReal}{\dataIN-F\xspace}
\newcommand{\dataCUBReal}{\dataCUB-F\xspace}
\newcommand{\dataInatReal}{\dataInat-F\xspace}
\newcommand{\dataCUBSyn}{\dataCUB-Rand\xspace}
\newcommand{\dataAirSyn}{\dataAir-Rand\xspace}

\newcommand{\methodHRN}{HRN\xspace}
\newcommand{\methodHCAST}{H-CAST\xspace}
\newcommand{\methodSSL}{Taxon-SSL\xspace}

\newcommand{\methodTextFull}{Text-guided Pseudo Attributes\xspace}
\newcommand{\methodTextViT}{Text-Attr (H-ViT)\xspace}
\newcommand{\methodTextCAST}{Text-Attr (H-CAST)\xspace}

\usepackage{dsfont}
\usepackage{pifont}
\newcommand{\cmark}{\ding{51}}%
\newcommand{\greencmark}{\textcolor{darkgreen}{\cmark}}%
\newcommand{\xmark}{\ding{55}}%
\newcommand{\redxmark}{\textcolor{darkred}{\xmark}}%

\newcommand{\levelFG}{Fine-grained}
\newcommand{\levelMid}{Subordinate}
\newcommand{\levelCoarse}{Basic}

\definecolor{cvprblue}{rgb}{0.21,0.49,0.74}
\usepackage[pagebackref,breaklinks,colorlinks,allcolors=cvprblue]{hyperref}

\def\paperID{***} 
\def\confName{CVPR}
\def\confYear{2026}

\newcommand{\papertitle}{
Free-Grained Hierarchical Visual Recognition
}
\title{\papertitle}
\author{%
\setlength{\tabcolsep}{12pt}
\begin{tabular}{@{}ccc@{}}
Seulki Park$^1$&
Zilin Wang$^1$&
Stella X. Yu$^{1,2}$\\
\end{tabular}\\
\setlength{\tabcolsep}{18pt}
\begin{tabular}{@{}cc@{}}
$^1$University of Michigan& 
$^2$UC Berkeley \\
\end{tabular}\\
\texttt{\{seulki,zilinwan,stellayu\}@umich.edu}
}

\begin{document}
\maketitle
\etocdepthtag.toc{mtchapter}
\etocsettagdepth{mtchapter}{subsection}
\etocsettagdepth{mtappendix}{none}
\faketableofcontents

\begin{abstract}
Hierarchical image recognition seeks to predict class labels along a semantic taxonomy, from broad categories to specific ones, typically under the tidy assumption that every training image is fully annotated along its taxonomy path.  Reality is messier: A distant bird may be labeled only bird, while a clear close-up may justify bald eagle. 

We introduce free-grain training, where labels may appear at any level of the taxonomy and models must learn consistent hierarchical predictions from incomplete, mixed-granularity supervision. We build benchmark datasets with varying label granularity and show that existing hierarchical methods deteriorate sharply in this setting. To make up for missing supervision, we propose two simple solutions: One adds broad text-based supervision that captures visual attributes, and the other treats missing labels at specific taxonomy levels as a semi-supervised learning problem. 

We also study free-grained inference, where the model chooses how deep to predict, returning a reliable coarse label when a fine-grained one is uncertain. Together, our task, datasets, and methods move hierarchical recognition closer to the way labels arise in the real world
\footnotemark.
\end{abstract}

\section{Introduction}
\label{sec:intro}

\def\figFreeGrainLearning#1{
\begin{figure}[#1]\centering
\includegraphics[width=0.96\linewidth]{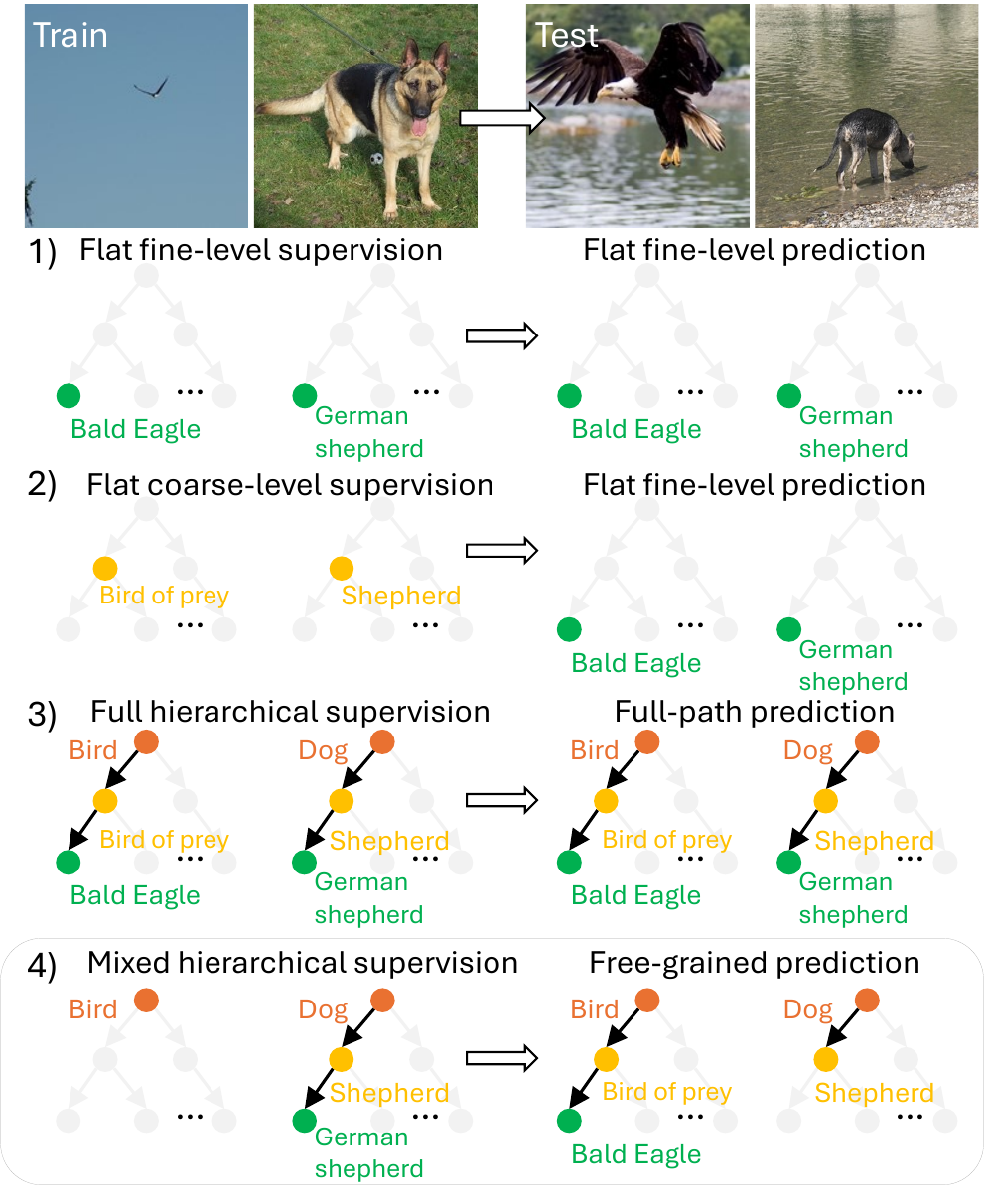}
\caption{ 
\textbf{We propose free-grained hierarchical recognition, where label granularity can freely vary across instances.} Both its training and inference differ from existing settings.
\textbf{1)} \textbf{Flat fine-grained recognition}: fine labels $\rightarrow$ fine predictions (fully supervised)~\cite{yun2024shvit}.
\textbf{2)} \textbf{Flat weakly supervised recognition}: coarse labels $\rightarrow$ fine predictions~\cite{falcon_2024}.
Both 1) and 2) operate on a single flat level without modeling cross-level relations.
\textbf{3)} \textbf{Hierarchical recognition}: full hierarchy $\rightarrow$ full hierarchy, requiring complete annotations at all levels~\cite{hcast2025park}.
\textbf{4)} \textbf{Free-grained recognition}: Labels may appear at different levels during training, e.g., a distant bird as \textit{Bird} and a close-up dog as \textit{German shepherd}. At inference, the model predicts at an appropriate level of specificity. The task thus requires learning from mixed, incomplete supervision and predicting hierarchical labels at the right depth.
\label{fig:freeGrainLearning}
}\vspace{-4mm}
\end{figure}
}

\figFreeGrainLearning{!t}
\setlength{\fboxsep}{0.5pt} 
\setlength{\fboxrule}{1.5pt}

\def\imw#1#2{\includegraphics[width=0.165\linewidth, height=0.12\linewidth]{#1}}

\def\rows#1#2{
\imw{figures/examples/#1/1.JPEG}{#2} &
\imw{figures/examples/#1/2.JPEG}{#2} &
\imw{figures/examples/#1/3.JPEG}{#2} &
\imw{figures/examples/#1/4.JPEG}{#2} &
\imw{figures/examples/#1/5.JPEG}{#2} &
\imw{figures/examples/#1/6.JPEG}{#2} &
\imw{figures/examples/#1/7.JPEG}{#2} 
}

\begin{figure*}[h]
\captionsetup[figure]{skip=2pt}
\small\centering
\newcommand{\colspace}{\hspace{2pt}}
\includegraphics[width=1\linewidth]
{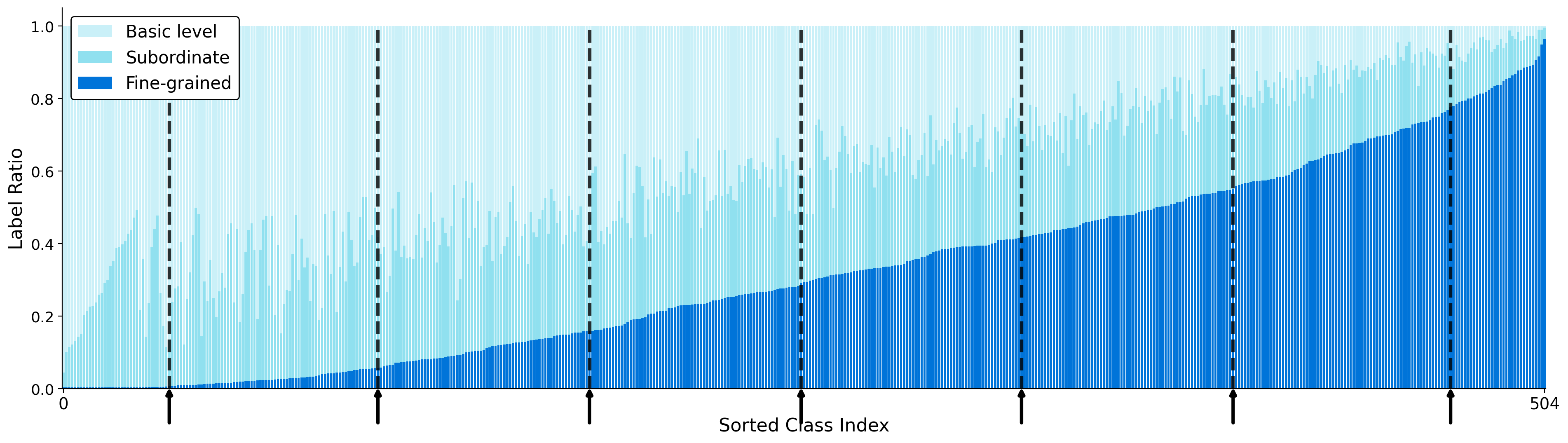} 
\resizebox{1\linewidth}{!}{
\hspace{0.4cm}
\begin{tabular}
{@{}
c@{\colspace}
c@{\colspace}c@{\colspace}c@{\colspace}
c@{\colspace}c@{\colspace}c@{}
}
 \fcolorbox{mybasic}{mybasic}{\makebox[2.75cm]{Dog}}  &  \fcolorbox{mybasic}{mybasic}{\makebox[2.75cm]{Cooking utensil}} & \fcolorbox{mybasic}{mybasic}{\makebox[2.75cm]{Dog}} &
 \fcolorbox{mybasic}{mybasic}{\makebox[2.75cm]{Fish}}  & \fcolorbox{mybasic}{mybasic}{\makebox[2.75cm]{Electron. equip.}} & \fcolorbox{mybasic}{mybasic}{\makebox[2.75cm]{Insect}}  & \fcolorbox{mybasic}{mybasic}{\makebox[2.75cm]{Bird}}  \\
\rows{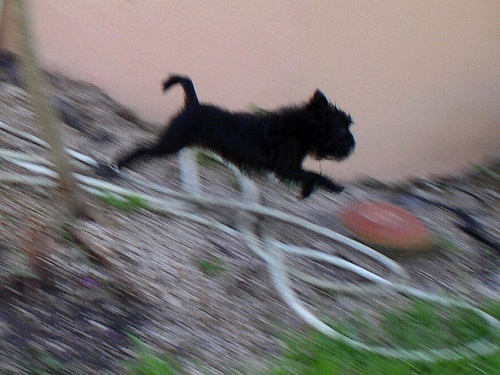}{mybasic}\\
\fcolorbox{mysub}{mysub}{\makebox[2.75cm]{Pinscher}}  & \fcolorbox{mysub}{mysub}{\makebox[2.75cm]{Pot}} & 
\fcolorbox{mysub}{mysub}{\makebox[2.75cm]{Shepherd}} &
\fcolorbox{mysub}{mysub}{\makebox[2.75cm]{Ganoid}} & \fcolorbox{mysub}{mysub}{\makebox[2.75cm]{Audio device}} & 
\fcolorbox{mysub}{mysub}{\makebox[2.75cm]{Beetle}} & \fcolorbox{mysub}{mysub}{\makebox[2.75cm]{Bird of prey}} \\
\rows{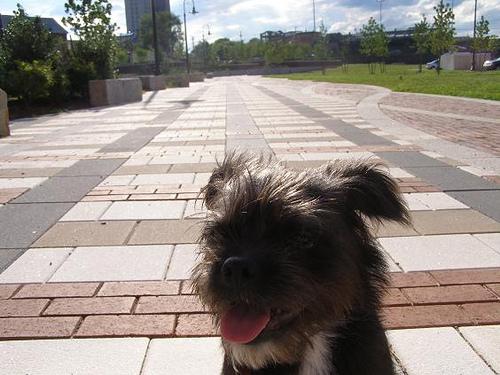}{mysub}\\
\fcolorbox{myfine}{myfine}{\makebox[2.75cm]{\textcolor{white}{Affenpinscher}}} 
& \fcolorbox{myfine}{myfine}{\makebox[2.75cm]{\textcolor{white}{Teapot}}}  
& \fcolorbox{myfine}{myfine}{\makebox[2.75cm]{\textcolor{white}{German shepherd}}}  &
\fcolorbox{myfine}{myfine}{\makebox[2.75cm]{\textcolor{white}{Gar}}}  & \fcolorbox{myfine}{myfine}{\makebox[2.75cm]{\textcolor{white}{Tape player}}}  & \fcolorbox{myfine}{myfine}{\makebox[2.75cm]{\textcolor{white}{Leaf beetle}}}  & \fcolorbox{myfine}{myfine}{\makebox[2.75cm]{\textcolor{white}{Bald eagle}}}  \\
\rows{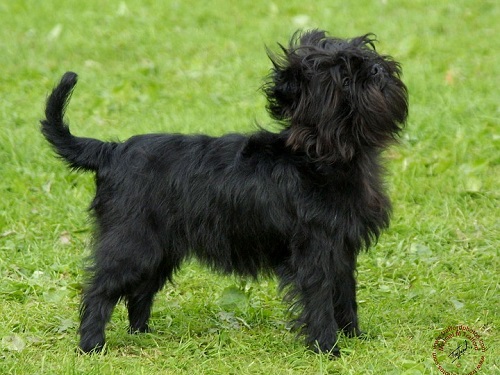}{myfine}\\
\end{tabular}
}
\vspace{-2mm}
\caption{\textbf{
Our \dataINReal reflects realistic mixed-granularity supervision, exhibiting both long-tailed fine-grained labels and visual ambiguity.}
{\bf Top:} Distribution of label depth (basic, subordinate, fine) across classes. Fine-grained labels are highly imbalanced, forming a long-tailed pattern where some classes retain many fine labels while others have few.
{\bf Bottom:} Examples illustrating confidence-based label assignment using foundation models. (Last column) A distant bird is labeled at the basic level (\colorbox{mybasic}{Bird}); a mid-range instance at the subordinate level (\colorbox{mysub}{Bird of prey}); and a clear close-up at the fine-grained level (\colorbox{myfine}{\textcolor{white}{Bald eagle}}).
}
\label{fig:label_proportion}
\vspace*{-2mm}
\end{figure*}

\footnotetext{Our dataset and code is available at \href{https://github.com/pseulki/FreeGrainLearning}{FreeGrainLearning}.}

Hierarchical classification~\citep{chang2021fgn,hrn2022chen,biderection2024jiang,hcast2025park} predicts a \textbf{semantic tree}  (\textit{Bird} $\to$ \textit{Bird of prey} $\to$ \textit{Bald eagle}), capturing categories from broad to specific. Predicting the full hierarchy can improve robustness and scalability, encouraging models to generalize across levels and making it easier to extend taxonomies with new parent or child classes. It also supports flexible use: An expert may want \textit{Bald eagle}, while a layman needs only \textit{Bird}.  Yet most existing methods~\citep{chang2021fgn,wang2023consistency} assume \textit{complete supervision}, where every training image is annotated along its full taxonomy path (Fig.~\ref{fig:freeGrainLearning}.\textcolor{refcolor}{3}).

Real annotations are rarely so tidy. Labels often appear at different levels of the taxonomy for two reasons. One is intrinsic: The image may not contain enough visual evidence to justify a fine-grained label. The other is extrinsic: annotation may be constrained by cost, expertise, or evolving labeling protocols~\cite{kim2023semanticcorrel,miao2023challenges}. Thus a distant image or non-expert annotator may provide only a coarse label such as \textit{Bird}, while a close-up or expert annotation may support a fine-grained label such as \textit{German shepherd}  (Fig.~\ref{fig:freeGrainLearning}.\textcolor{refcolor}{4}).

To capture this messier reality, we propose {\it free-grained hierarchical recognition}, where supervision is free to vary in granularity: training labels may appear at any level of a taxonomy, from coarse to fine, and may differ across instances. At inference, the model outputs a free-grained prediction, selecting the deepest label it can reliably support.

The challenge of free-grained learning is to learn from supervision that is {\it incomplete, uneven, and spread across levels}, while still producing predictions that remain consistent with the taxonomy. This setting differs from conventional ones, where labels are provided at a single level -- either fine-grained in the fully supervised case or coarse in the weakly supervised case -- without requiring the model to connect information across levels (Fig.~\ref{fig:freeGrainLearning}.\textcolor{refcolor}{1-2}).

To support this setting, we construct new benchmarks by adapting existing hierarchical datasets (\dataCUB~\cite{ref:data_cub200}, \dataAir~\cite{ref:data_air}, \dataInat~\cite{inat21mini}) to exhibit mixed-granularity supervision. 
While these benchmarks provide valuable testbeds, they are limited in scale or diversity: \dataCUB and \dataAir are small-scale, and \dataInat is confined to a single biological domain. Larger datasets such as \dataIN~\cite{ref:data_imagenet} inherit deep, inconsistent hierarchies from WordNet~\cite{fellbaum1998wordnet}, and are therefore  rarely used for prior hierarchical recognition~\cite{chang2021fgn, hrn2022chen, hcast2025park}. 
We thus redesign \dataIN into a clean three-level tree for hierarchical recognition. 

On top of these datasets, we construct two complementary variants to capture realistic annotation difficulty and varying label availability. {\bf 1) Foundation-based variants} (\dataINReal, \dataInatReal, \dataCUBReal) set label depth by whether large visual foundation models~\cite{ref:clip_2021, bioclip_2024_CVPR} correctly predict each level. Though imperfect as a proxy for human annotation, they provide a practical approximation: Deeper-level errors often coincide with visual ambiguity, yielding realistic mixed-granularity supervision. Thus distant birds may be labeled as \textit{Bird}, mid-range ones as \textit{Bird of prey}, and clear close-ups as \textit{Bald eagle} (Fig.~\ref{fig:label_proportion}). This also induces long-tailed label availability, since fine-grained labels are removed more often for some classes than for others. {\bf 2) Randomized variants} (\dataCUBSyn, \dataAirSyn) assign label depths at varying proportions, enabling systematic evaluation under different levels of label availability. Together, these variants span diverse mixed-granularity scenarios and pose a broad benchmark for free-grain learning.

Our setting is hard for existing hierarchical classifiers. When trained under free-grained supervision, state-of-the-art methods~\citep{hrn2022chen, hcast2025park} lose up to \textbf{40\%} in full-path accuracy on \dataInat~\cite{inat21mini}, where a prediction counts as correct only if every level of the taxonomy is correct. This sharp deterioration underscores the difficulty of learning from mixed-granularity labels and the need for methods that can handle incomplete supervision more robustly.

To address this, we propose two {\it free-grained training} methods that compensate for missing supervision in different ways. {\bf 1) Text-guided pseudo-attributes} add auxiliary text supervision in the form of image descriptions generated by a vision–language model~\cite{dubey2024llama3}, providing semantic cues about visual attributes that help the model learn discriminative features even when fine-grained labels are absent. {\bf 2) Taxonomy-guided semi-supervised learning} (SSL) instead treats missing labels at particular taxonomy levels as unlabeled data, using hierarchical consistency to learn from both labeled and unlabeled examples. Across datasets, each method improves over state-of-the-art hierarchical classifiers by 5--25\%, providing strong baselines while also underscoring the remaining difficulty of free-grained learning.

We also study {\it free-grained inference}, in which the model adaptively chooses how deep to predict. This is motivated by a simple practical point: A correct coarse prediction is often preferable to an incorrect fine-grained one. We consider two strategies: {\bf 1) confidence-based}, selecting the deepest label with sufficient confidence, and {\bf 2) consistency-based}, selecting the deepest level that maintains hierarchical consistency.  We find that the latter yields more reliable and deeper correct predictions.


{\bf Contributions.}
{\bf 1)} We introduce free-grained hierarchical visual recognition, where training labels may appear at any taxonomy level and inference adaptively chooses prediction depth.
{\bf 2)} We build benchmark datasets for this setting, including foundation-based variants for realistic annotation difficulty and randomized variants for controlled label availability.
{\bf 3)} We propose text-guided pseudo-attributes and taxonomy-guided SSL, both of which outperform prior hierarchical classifiers under mixed-granularity supervision.
{\bf 4)} We show that, for free-grained inference, consistency-based inference yields more reliable and deeper correct predictions than confidence-based inference.
\section{Related Work}

\textbf{Hierarchical classification} predicts the full taxonomy path for each image, requiring {accurate level-wise predictions} while also encouraging {parent–child consistency} across the hierarchy~\cite{chang2021fgn, hrn2022chen, wang2023consistency, hcast2025park}.
Meanwhile, some methods use the hierarchy as an auxiliary signal for flat (fine-grained) classification, for example, to regularize feature learning~\cite{zhang2022use} or to reduce the severity of fine-grained mistakes~\cite{karthik2021nocost, garg2022mistake}.
Importantly, all these approaches \textit{assume complete hierarchical supervision} is available for every training sample.

\noindent\textbf{Long-tailed and semi-/weakly-supervised recognition} address distinct real-world challenges~\citep{LT_liu_2019_CVPR, LT_Park_2022_CVPR, wu2023chmatch, wsrobinson20a},  typically operating at a \emph{single} granularity, using either fine-grained labels alone or coarse labels alone.
In contrast, our {free-grained learning} introduces a new, unexplored problem: learning from \emph{mixed-granularity} labels within a hierarchy.
This naturally brings together challenges from multiple areas, including class imbalance within each level, imbalance across different levels of the hierarchy, weak/semi-supervision, and the need to maintain hierarchical consistency, all within a single unified framework.
Unlike prior work that addresses these challenges in isolation, our setting requires handling them jointly under mixed-granularity supervision (Table~\ref{table:task_comparison}).
See more related works in Appendix~\ref{sec:suppl_related_work}.

\newcommand{\greenmark}{\textcolor{textgreen}{\ding{51}}}
\newcommand{\redmark}{\textcolor{darkred}{\ding{55}}}

\begin{table}[h]
\centering 
\caption{\textbf{Our task is more practical and challenging.}
Free-grained learning reflects real-world annotation, where each image may have fine (F) or coarse (C) labels, and models must predict a taxonomy-consistent hierarchy.
It jointly introduces class imbalance (Cls. Imb.) and level imbalance (Lvl. Imb.), along with weak and partial supervision—factors mostly studied in isolation.
Evaluation considers both accuracy (Acc.) and consistency (Con.).
}\label{table:task_comparison}
\vspace{-2mm}
\resizebox{1\linewidth}{!}{
{\footnotesize
\setlength{\tabcolsep}{2pt}
\begin{tabular}{@{}l|cc|cc|c|cc|cc@{}}
\toprule
& \multicolumn{2}{c|}{\textbf{Input}}                            & \multicolumn{2}{c|}{\textbf{Output}}                           & \textbf{Labels} & \multicolumn{2}{c|}{\textbf{Imbalance}}                        & \multicolumn{2}{c}{\textbf{Metrics}}                \\ \cline{2-10} 
\textbf{Tasks}                      & F.                      & C.                    & F.                      & C.                    & Avail.    & Cls.                     & Lvl.                     & Acc.                  & Con.               \\ \hline
Long-tailed recog.   & \greenmark &      \redmark                    & \greenmark &     \redmark & All             & \greenmark &      \redmark                      & \greenmark &   \redmark                         \\ \hline
Semi-supervised   & \greenmark &   \redmark                        & \greenmark &    \redmark                       & \textit{Partial}         &     \redmark                      &    \redmark                       & \greenmark &      \redmark                     \\ \hline
Weakly-supervised &      \redmark                     & \greenmark & \greenmark &        \redmark                   & All             &       \redmark                    &   \redmark                        & \greenmark &          \redmark                 \\ \hline
Hierarchical recog.  & \greenmark & \greenmark & \greenmark & \greenmark & All             &     \redmark                      &      \redmark                     & \greenmark & \greenmark \\ \hline
\rowcolor{suborange!20} 
\textbf{Free-grained recog.}   & \greenmark & \greenmark & \greenmark & \greenmark & \textit{Partial}         & \greenmark & \greenmark & \greenmark & \greenmark \\ \bottomrule
\end{tabular}
}
}
\vspace{-3mm}
\end{table}

\begin{table}[b]\centering
\caption{
\textbf{We convert existing hierarchical benchmarks into free-grain versions.}
Since ImageNet’s taxonomy is inconsistent, we newly curate a consistent three-level hierarchy, \dataIN-3L.
}\vspace{-3mm}
\label{table:dataset_comparison}
\setlength{\tabcolsep}{3pt}
\small
\resizebox{1\linewidth}{!}{
\begin{tabular}{l|c|r|r|r}
\toprule
 \textbf{Dataset}              & \textbf{\#levels} & 
 \textbf{\#classes per level} & \textbf{\#train} & \textbf{\#test} \\ \hline
CUB            & 3                & 13-38-200                  & 5,994       & 5,794   \\ \hline
\dataAir      & 3                & 30-70-100                  & 6,667       & 3,333   \\ \hline
\dataInat      & 8                & 3-11-13-51-273-1103-4884-10000                  & 500,000       & 100,000   \\ \hline
\dataIN       & 5-19             & - 1000                         & 1,281,167      & 50,000  \\ \hline
\rowcolor{suborange!20} 
\textbf{\dataIN-3L}    & \textbf{3}                & \textbf{20-127-505}                 & \textbf{645,480}        & \textbf{25,250}  \\ 
\bottomrule
\end{tabular}
}\vspace{-5mm}
\end{table}



\begin{figure}[!h]
\centering
\makebox[\linewidth]{%
\hspace*{0.02\linewidth}
\begin{subfigure}[t]{0.64\linewidth}
    \includegraphics[height=3.5cm]{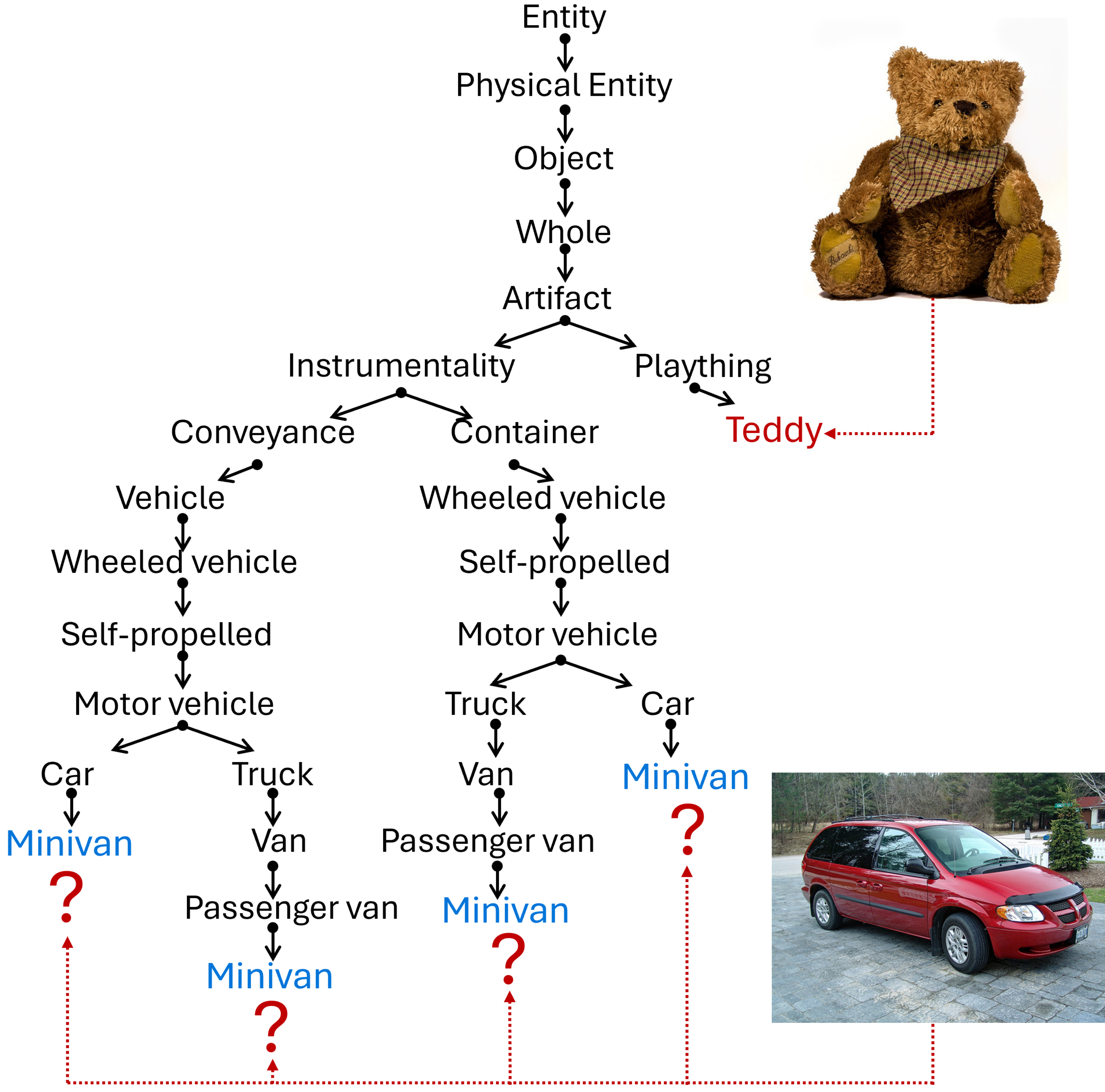}
    \subcaption{\small Sample hierarchies in ImageNet}
    \label{fig:minivan_teddy2}
\end{subfigure}
\hfill
\begin{subfigure}[t]{0.35\linewidth}
    \includegraphics[height=3.5cm]{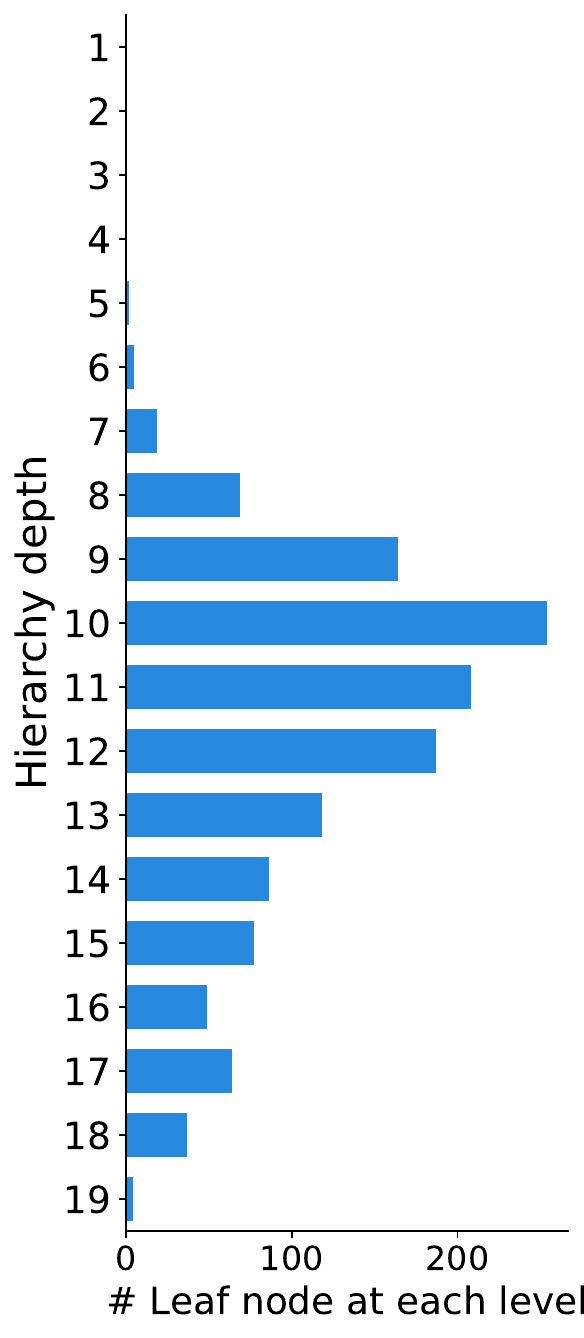}
    \subcaption{\small Hierarchy depths}
    \label{fig:depth_hist}
\end{subfigure}
}
\makebox[\linewidth]{%
\hspace*{0.02\linewidth}
\begin{subfigure}[t]{0.38\linewidth}
    \includegraphics[width=\linewidth,height=3.3cm]{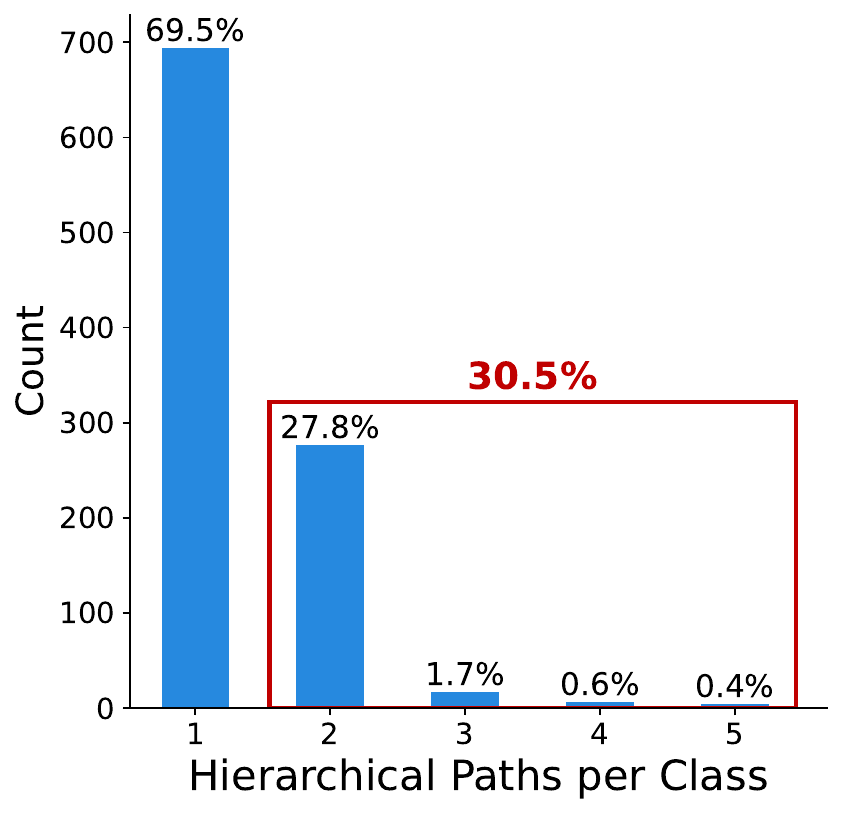}
    \subcaption{\small \# paths per class}
    \label{fig:minivan_teddy2}
\end{subfigure}
\hfill
\begin{subfigure}[t]{0.6\linewidth}
    \makebox[\linewidth][r]{%
    \includegraphics[width=\linewidth, keepaspectratio,height=3.3cm]{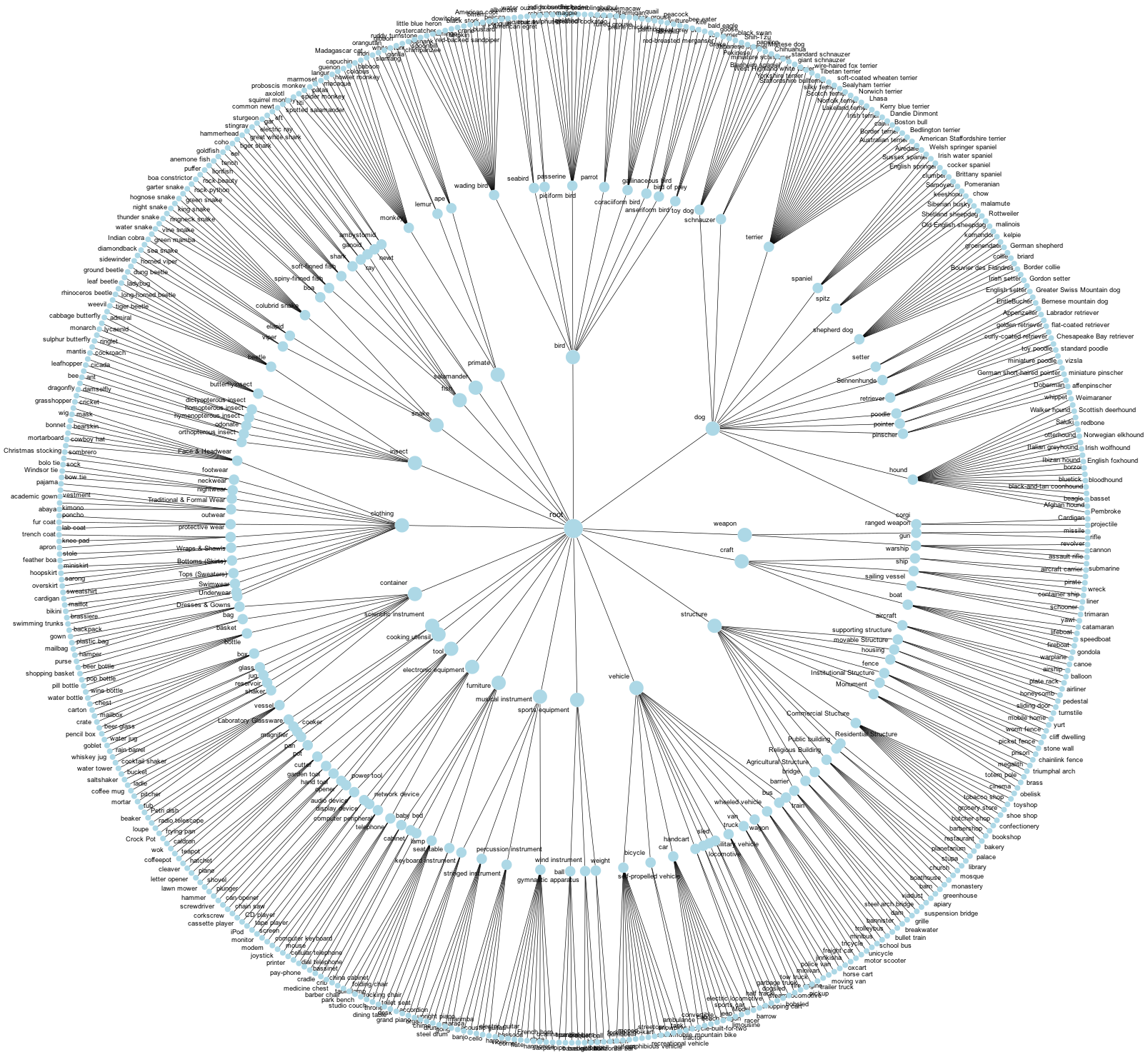}\hspace*{0.1\linewidth}
    }
    \subcaption{\small Our 3-level hierarchy}
    \label{fig:depth_hist}
\end{subfigure}
}
\vspace{-5pt}
\caption{
{\bf We curate \dataIN-3L as a benchmark for hierarchical classification.}
{\bf (a)} Sample hierarchies reveal two issues: 1) some classes have multiple valid paths (e.g., \textit{Minivan}), and 2) classes at the same depth can have mismatched specificity (e.g., \textit{Teddy} vs. \textit{Conveyance}).  
{\bf (b)} ImageNet classes span widely varying depths (5–19 levels), often exceeding 10, highlighting inconsistency in hierarchy depth.  
{\bf (c)} 30\% of classes have multiple valid hierarchical paths, introducing ambiguity in evaluation.  
{\bf (d)} We construct a coherent 3-level taxonomy, inspired by cognitive psychology~\citep{rosch1976basic}: {\it basic} for general recognition, {\it subordinate} for contextual specificity, and {\it fine-grained} for specialized distinctions.
}
\vspace*{-12pt}
\label{fig:imagenet_taxonomy}
\end{figure}

\section{Benchmarks for Free-Grained Recognition}\label{sec:dataset}
We adapt existing hierarchical benchmarks to our free-grained setting, but prior datasets are often small-scale (e.g., \dataCUB, \dataAir) or domain-specific (e.g., \dataInat), as shown in Table~\ref{table:dataset_comparison}.
To enable a large-scale and diverse benchmark, we reorganize \dataIN into a clean three-level hierarchy (\dataIN-3L), as its original WordNet taxonomy is irregular (Fig.~\ref{fig:imagenet_taxonomy}). This simplified structure supports mixed-granularity prediction without unnecessary complexity.
We first describe this restructuring (Sec.~\ref{subsec:3.1}), then introduce foundation-based variants that mimic real-world annotation patterns (Sec.~\ref{subsec:3.2}) and randomized variants for controlled evaluation (Sec.~\ref{subsec:3.3}).

\subsection{\bf Constructing \dataIN-3L}\label{subsec:3.1}
As in Fig.~\ref{fig:imagenet_taxonomy}, the WordNet hierarchy~\cite{fellbaum1998wordnet} is noisy and inconsistent, making ImageNet unsuitable for full-path evaluation.
To address this, we simplify the WordNet hierarchy by removing overly abstract nodes (e.g., \textit{Entity}, \textit{Whole}) and restructuring it into a consistent three-level taxonomy. This design is guided by categorization principles~\citep{rosch1976basic}, where the \emph{basic} level is the most natural and visually distinctive. 

We anchor categories at a shared \emph{basic} level (e.g., \emph{vehicle, dog}) and organize subordinate and fine-grained categories under it. When multiple candidates exist, we select those that best match the granularity of existing basic classes. We further apply additional design principles, described below, with full details provided in the appendix~\ref{sec:suppl_data_construction}.

\noindent\textbf{1) \textit{Enforce meaningful structure:}}
We remove paths where each node has only one child, since coarse labels fully determine the fine labels. Branches with fewer than three levels are also excluded.
\textbf{2) \textit{Maximize within-group diversity:}}
Among subordinate candidates under each basic class, we favor those with richer fine-grained subclasses, for example choosing \textit{parrot} (4 children) over \textit{cockatoo} (1 child).
\textbf{3) \textit{Refine vague categories:}} 
Ambiguous groups such as \textit{Women’s Clothing} are reorganized into precise, functionally grounded categories (e.g., \textit{Underwear}) to improve clarity.
\textbf{4) \textit{Validate with language models and human review:}}
We use large language models (ChatGPT~\citep{achiam2023gpt4}) to suggest refinements, with all decisions manually reviewed for semantic consistency.
Applying this curation process to ImageNet-1k yields a structured benchmark of 20 basic, 127 subordinate, and 505 fine-grained classes, \dataIN-3L, ensuring every branch supports meaningful hierarchical prediction (a complete list is provided in Appendix~\ref{sec:suppl_complete_hierarchy}).

\begin{figure*}[!t]
    \centering
    \subfloat[\textbf{Text-Attr}]{
        \includegraphics[width=0.44\linewidth]{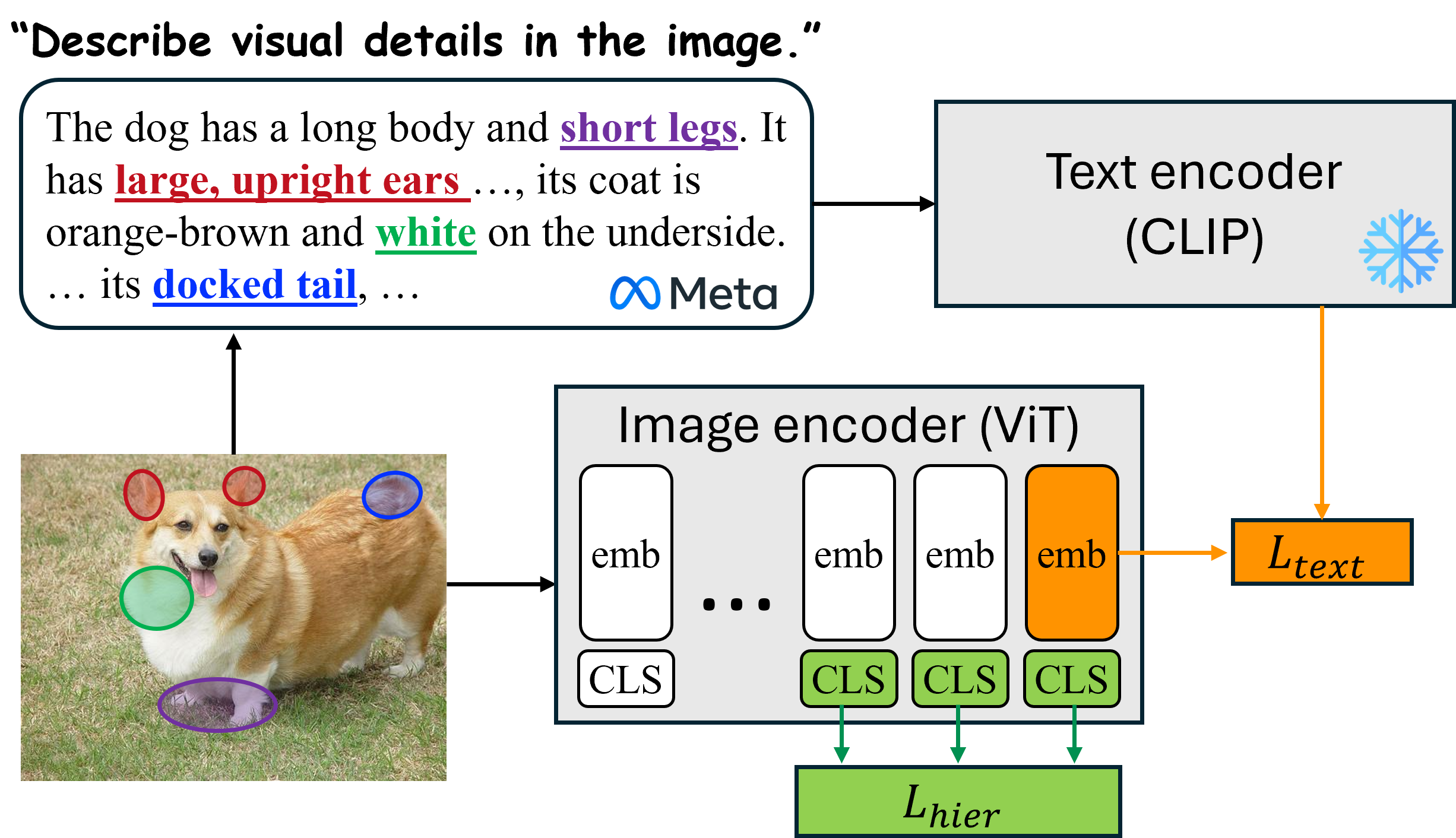}
        \label{fig:method_text}
    }
    \hspace{0.03\linewidth}
    \subfloat[\textbf{\methodSSL}]{
        \includegraphics[width=0.44\linewidth]{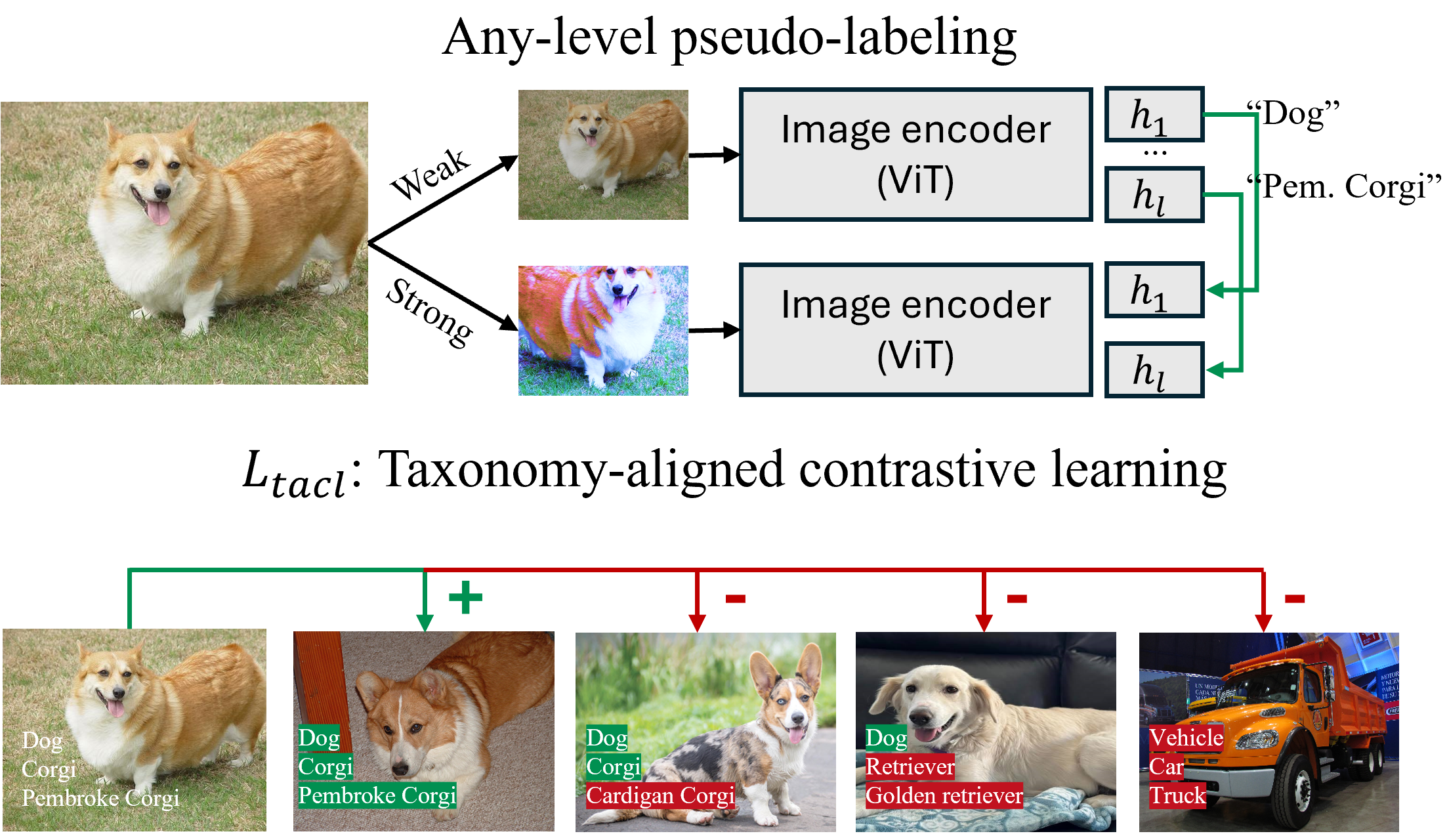}
        \label{fig:method_ssl}
    }\vspace{-2mm}
    \caption{\textbf{Overview of the proposed methods.} 
    (a) Text-Attr enriches feature representations using semantic cues from images, compensating for missing labels and capturing shared attributes across levels.
    (b) \methodSSL handles missing-level labels by treating them as unlabeled and learns from visual consistency through augmented views.
    Both methods offer distinct benefits for our challenging task. }\vspace{-3mm}
    \label{fig:method_overview}
\end{figure*}

\subsection{Foundation-based Pruning}\label{subsec:3.2}
To build realistic free-grain training sets, we prune hierarchical labels using large vision–language models: CLIP~\citep{ref:clip_2021} for \dataINReal and BioCLIP~\citep{bioclip_2024_CVPR} for \dataInatReal and \dataCUBReal.
While these models are not designed to measure ambiguity, their zero-shot confidences indicate when fine-grained labels are less reliable, whether due to limited visual detail, annotation difficulty, or inconsistent expertise.
Fig.~\ref{fig:label_proportion} shows that this results in mixed-granularity supervision patterns that often follow such visual ambiguity, with distant or less discernible instances labeled more coarsely and clearer ones labeled more finely.

We adopt CLIP’s prompt-ensemble strategy (e.g., \textit{a photo of a [class]}) and assign labels from coarse to fine based on prediction correctness:
\textbf{(1)} We always retain the \emph{basic} label.
\textbf{(2)} If the subordinate prediction is correct, we retain the \emph{subordinate} label.
\textbf{(3)} If both subordinate and fine-grained predictions are correct, we retain the \emph{fine-grained} label.
This defines the deepest available label for each image, with higher-level labels assumed from the given taxonomy.
Since relying solely on foundation models' predictions can produce biased label distributions, we further remove a portion of subordinate labels based on the fine-grained removal rate per class, introducing more challenging supervision.

This pruning only removes labels and introduces no additional semantic information; it may even increase difficulty by discarding labels of harder examples. While we use CLIP and BioCLIP, similar pruning can be performed with other foundation models or ensembles (e.g., removing labels consistently mispredicted across models).

\noindent\textit{\bf 1) \dataINReal.} After pruning, 32.6\% of images retain all three levels (Basic + Subordinate + Fine-grained), 28.0\% retain two (Basic + Subordinate), and 39.4\% retain only the Basic. Each class keeps the same number of images as in ImageNet; imbalance arises only from label granularity.

\noindent\textit{\bf 2) \dataInatReal.} BioCLIP, a biology foundation model, performs well on species-level prediction but struggles with coarser labels. This mismatch enables substantial pruning: 22.5\% of images retain all three levels (Order + Family + Species), 28.0\% retain two, and 49.5\% retain only Order.

\noindent\textit{\bf 3) \dataCUBReal.} With the same procedure, 31.5\% of images keep 3 levels, 23.3\% two (Order, Family), 45.2\% only Order.

\subsection{Randomized Pruning}\label{subsec:3.3}
To control label availability, we construct randomized variants, \dataCUBSyn and \dataAirSyn, by randomly pruning labels from \dataCUB~\citep{ref:data_cub200} and \dataAir~\citep{ref:data_air}. Unlike realistic pruning, this design systematically varies supervision and simulates \textit{extreme} sparsity (e.g., only 10\% fine-grained labels), enabling stress-testing of model robustness across diverse label distributions. Although random removal is independent of image difficulty, it reflects practical factors such as annotator expertise, cost, or task-specific constraints.
We denote availability as $a$-$b$-$c$, where $a$\% of basic, $b$\% of subordinate, and $c$\% of fine-grained labels are retained (e.g., 100-50-10 retains 10\% fine-grained labels and 40\% subordinate-only labels).

\section{Free-Grain Learning Methods}

We define our free-grained hierarchical recognition problem and introduce two ways to handle missing supervision.


\subsection{Problem Setup}\label{subsec:problem}
In free-grained hierarchical classification, the goal is to predict labels across all levels of a taxonomy from training data with mixed granularity. Each image is annotated at a certain level, and all coarser labels are assumed to be available while finer ones are missing; the coarsest label is always given. The model is trained to produce consistent predictions across the full hierarchy.

\noindent\textbf{Free-grained Hierarchical Loss.}
To adapt prior hierarchical recognition methods to the free-grained setting, we modify their hierarchical supervision by applying the loss only at levels with available labels.
Given hierarchical labels $y_1, \dots, y_L$ across $L$ levels, the loss is defined as:
\vspace{-1mm}
\begin{equation} \label{eq2:loss_hier}
    \mathcal{L}_{\text{hier}} = \sum_{l=1}^{L} \mathbbm{1}_{\{y_l \text{ exists}\}} \cdot \mathcal{L}( f_l(x), y_l ),
\end{equation}
where $f_l(x)$ is the prediction at level $l$, and $\mathcal{L}$ denotes a classification loss (e.g., cross-entropy).

\subsection{Semantic Guidance with Text Attributes}
\label{method1:text}
Our semantic guidance approach is motivated by the observation that while class labels differ across hierarchical levels (e.g., \textit{Dog} → \textit{Corgi} → \textit{Pembroke}), many visual attributes, such as ``tail length'' or ``ear shape'', remain consistent (Fig.~\ref{fig:method_text}). To capture these shared semantic cues, we use image descriptions as auxiliary supervision. 
While recent large language models (LLM) (e.g., ChatGPT~\cite{achiam2023gpt4})-based approaches such as FineR~\cite{liu2024democratizing} also use vision-language model (VLM)-generated text, their purpose is different: they feed these cues into an LLM for training-free fine-grained class reasoning, whereas we use text as supervision to train image representations to capture visual attributes shared across hierarchical levels.

Specifically, given an input image $x$, we use a frozen vision-language model (VLM), Llama-3.2-11B~\citep{dubey2024llama3}, to generate a language description $d_x$, using the prompt:
``\textit{Describe visual details in the image.}"
This produces descriptions containing phrases such as ``short legs” or ``pointed ears,” which we encode into a text embedding $z_x^t$ using CLIP’s text encoder~\citep{ref:clip_2021}. We cap generation at 100 tokens, while CLIP accepts 77 tokens; longer descriptions are truncated during encoding. Although truncation discards some details, our method focuses on shared semantic cues (e.g., “short legs,” “brown markings”) rather than exhaustive captions, making it robust to this limitation.
In parallel, we obtain the image embedding $z_x^v$ from the image encoder, and align it with the text embedding $z_x^t$ using a contrastive loss:
\begin{equation}
    \mathcal{L}_{\text{text}} = -\frac{1}{N} \sum_{i=1}^{N} \log \left( \frac{\exp(\text{sim}(z_i^v, z_i^t) / \tau)}{\sum_{j=1}^{N} \exp(\text{sim}(z_i^v, z_j^t) / \tau)} \right),
\end{equation}
where $\text{sim}(\cdot,\cdot)$ is cosine similarity and $\tau$ is a temperature parameter.
This loss guides the encoder to capture salient, label-independent traits shared across levels. Although not explicitly predicting attributes, aligning image features with text induces intermediate representations, which we call {pseudo-attributes}. This model-agnostic method can be applied to any architecture.

\subsection{Visual Guidance with SSL}
\label{method2:semi}

We adopt a semi-supervised formulation because missing-grain labels can be treated as unlabeled data.
CHMatch~\citep{wu2023chmatch} shows that coarse labels can improve pseudo-labeling, but it is limited to a two-level (coarse–fine) setting and focuses on refining fine-grained predictions.
We generalize this to arbitrary multi-level taxonomies by 1) generating pseudo-labels at \emph{every level}, and 2) enforcing \emph{cross-level consistency} so that predictions remain valid along the hierarchy.

\noindent
\textbf{1) Multi-level pseudo-labeling.}
Following CHMatch, we decouple the classifier \( f \) into a shared feature extractor \( f_{\text{feat}} \) and level-specific heads \( \{h_l\}_{l \in \mathcal{S}_x} \), where each head predicts labels at a different taxonomy level.
The supervised loss is computed using Eq.~\ref{eq2:loss_hier}, applying supervision only at levels with available labels.
Pseudo-labels at each level are generated from the predictions of the corresponding head given a weakly augmented input \( W(x) \).

\noindent
\textbf{2) Taxonomy-aligned feature learning.}
A key challenge is that pseudo-labels at different levels may be inconsistent (e.g., two samples share a coarse label but differ at fine levels).
To address this, we only treat pairs as \emph{reliable positives} when they agree across \emph{all levels}.

For each mini-batch, we build level-wise affinity graphs $W^l$ based on pseudo-label agreement: $W^l_{ij}=1$ if images $i$ and $j$ share the same pseudo-label at level $l$, and 0 otherwise. 
We then define a taxonomy-aligned affinity:
\begin{equation}
    W_{ij} =
    \begin{cases}
        1 & \text{if } W^1_{ij} = \cdots = W^L_{ij} = 1, \\
        0 & \text{otherwise.}
    \end{cases}
\end{equation}

\noindent
This enforces that two samples are considered similar \emph{only if they are consistent across the entire hierarchy}, effectively filtering out noisy or partially incorrect pseudo-labels.

\noindent
\textbf{Contrastive objective.}
Using \(W\), we pull together positive pairs (\(W_{ij}=1\)) and push apart negative pairs (\(W_{ij}=0\)).
The taxonomy-aligned contrastive loss is:
\begin{equation}
\footnotesize
\mathcal{L}_{\text{tacl}} =
-\frac{1}{\sum_{j}W_{ij}}
\log
\frac{
\sum_{j} W_{ij}\exp\big((g(f(x_i)) \cdot g(f(x_j))')/t\big)
}{
\sum_{j} (1-W_{ij})\exp\big((g(f(x_i)) \cdot g(f(x_j))')/t\big)
},
\end{equation}
where \(g(f(x_i))\) is the projected feature of image \(i\), and \(t\) is a temperature.
This objective encourages samples with consistent hierarchical semantics to form tight clusters in the feature space, while separating those that disagree at any level.
See more details in the Appendix~\ref{supple:sec:taxon-ssl}.

\begin{figure*}[t]
\centering
\begin{subfigure}{0.23\linewidth}
    \centering
    \includegraphics[width=\linewidth]{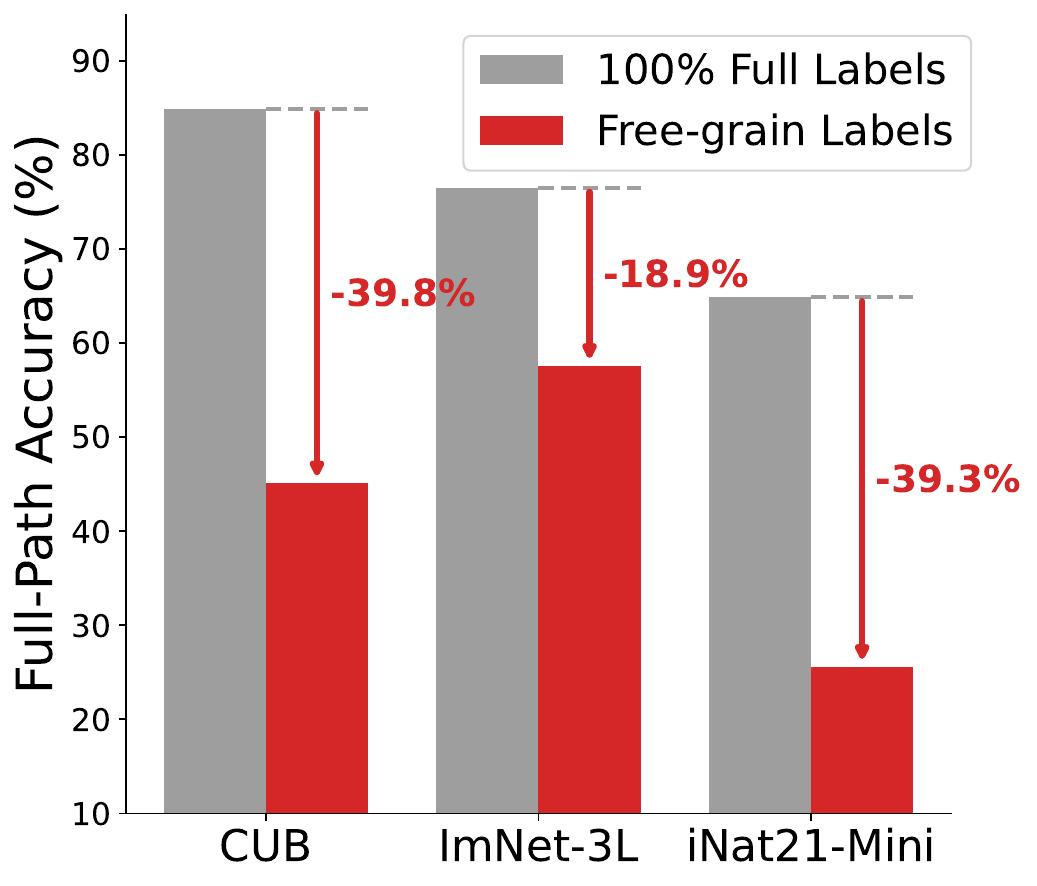}
    \vspace{-5mm}
     \subcaption{\bf H-CAST: full vs. free-grain}
    \label{fig:performance_drop}
\end{subfigure}
\hfill
\begin{subfigure}{0.75\linewidth}
    \centering
    \resizebox{\linewidth}{!}{%
    \begin{tabular}{l >{\centering\arraybackslash}p{0.9cm}>{\centering\arraybackslash}p{0.9cm}>{\centering\arraybackslash}p{0.9cm}>{\centering\arraybackslash}p{0.9cm}>{\centering\arraybackslash}p{0.9cm}|>{\centering\arraybackslash}p{0.9cm}>{\centering\arraybackslash}p{0.9cm}>{\centering\arraybackslash}p{0.9cm}>{\centering\arraybackslash}p{0.9cm}>{\centering\arraybackslash}p{0.9cm}}  
\toprule
Dataset & \multicolumn{5}{c|}{\dataINReal (20-127-505)}             & \multicolumn{5}{c}{\dataInatReal (273 - 1,103 - 10,000)}               \\ \cline{2-11} 
        & FPA($\uparrow$)   & fine.($\uparrow$) & sub.($\uparrow$) & basic($\uparrow$)  & {\small TICE($\downarrow$)}  & FPA($\uparrow$)   & spec.($\uparrow$) & fam.($\uparrow$) & order($\uparrow$)  & {\small TICE($\downarrow$)}  \\ \midrule
\methodHRN~\citep{hrn2022chen}           & 37.79 & 38.73   & 55.73 & 78.65     & 46.69  & 17.03 & 25.43   & 46.51 & 70.20     & 53.81\\
\methodHCAST~\citep{hcast2025park}& \underline{57.59} & 59.02   & \underline{82.69} & \underline{93.53}     & 21.81 & 25.63 & 28.61   & 67.20 & 83.62     & 47.17 \\ \midrule
\methodSSL & 48.40 & 52.34   & 65.74 & 82.96     & 19.87 & \underline{31.74} & \textbf{37.11}   & 69.53 & 82.02    & \underline{37.31}\\ 
\methodSSL + Text-Attr& 49.65 & 53.43   & 66.43 & 83.56     & \underline{18.81} & \textbf{31.93} & \underline{37.08}   & \underline{69.76} & \underline{82.20}     & \textbf{37.04}\\  
\noalign{\vskip 2pt}
\hdashline
\noalign{\vskip 2pt}
\methodTextViT          & 55.48 & \underline{59.05}   & 77.95 & 89.45     & 24.02  & 27.88 & 32.07   & 68.27 & 80.49     & 46.35\\
\methodTextCAST & \textbf{63.20} & \textbf{64.91}   &\textbf{ 84.47} & \textbf{93.56}     & \textbf{18.58} &29.74 & 32.37   & \textbf{71.79} & \textbf{85.99}     & 44.63\\ \bottomrule
\end{tabular}
}
    \subcaption{\bf Method comparison on \dataINReal and \dataInatReal.}
    \label{tab:main_imagenet_real}
\end{subfigure}
\vspace{-2mm}
\caption{\textbf{(a) Transitioning from fully labeled data to our free-grain setting results in a substantial drop in Full-Path Accuracy, highlighting the difficulty of the task.} SOTA H-CAST drops by 19--40 percentage points across datasets.  
\textbf{(b) Our methods effectively improve performance under free-grain supervision, with behavior depending on data characteristics.} Conventional hierarchical methods such as \methodHRN~\citep{hrn2022chen} and \methodHCAST~\citep{hcast2025park} degrade significantly under incomplete supervision. In contrast, \methodTextCAST performs strongly on \dataINReal, where rich visual cues support text-guided learning, while \methodSSL is more effective on \dataInatReal, where fine-grained classes have similar appearances. Combining both (\methodSSL + Text-Attr) yields consistent but modest gains across datasets.}
\label{fig:main_result}
\vspace{-5mm}
\end{figure*}

\section{Experiments}
In this section, we first describe the experimental setup (Sec.~\ref{subsec:exp_setup}). We then present results on our free-grain benchmarks (Sec.~\ref{subsec:results_benchmark}) and provide further analysis of the proposed methods (Sec.~\ref{subsec:analysis}). Finally, we compare free-grained inference methods (Sec.~\ref{subsec:free-inference}).

\subsection{Experimental Setup}\label{subsec:exp_setup}

\noindent\textbf{1) Dataset:}
We conduct experiments using our proposed \textit{\dataINReal}, \textit{\dataInatReal}, and \textit{\dataCUBReal} datasets, along with the synthetic \textit{\dataCUBSyn} and \textit{\dataAirSyn} datasets.

\vspace{2pt}
\noindent\textbf{2) Evaluation metrics:}
Following~\citep{hcast2025park}, we evaluate accuracy and consistency:
\textit{(1) Level-accuracy}: Top-1 accuracy at each level.  
\textit{(2) Tree-based InConsistency Error rate (TICE)}: Proportion of samples with inconsistent predictions in the hierarchy (lower is better):
$\text{TICE} = \frac{n_{\text{ic}}}{N}$, where $N$ is the total number of samples and $n_{\text{ic}}$ is the number of inconsistent predictions.  
\textit{(3) Full-Path Accuracy (FPA)}: Proportion of samples correctly predicted at all levels (primary metric): 
$\text{FPA} = \tfrac{n_{\text{ac}}}{N}$, where $n_{\text{ac}}$ is the number of samples correct at all hierarchy levels.

\vspace{2pt}
\noindent\textbf{3) Comparison Methods:}
We adapt two strong and relevant hierarchical classifiers to the free-grained setting for comparison.
\textit{(1) Hierarchical Residual Network (\methodHRN)}~\citep{hrn2022chen}: the first to handle supervision at both subordinate and fine-grained levels by maximizing marginal probabilities within the tree-constrained space.
\textit{(2) \methodHCAST}~\citep{hcast2025park}: the current SOTA, encouraging consistent visual grouping across taxonomy levels. Originally trained with full supervision, we adapt it to this setting via the level-wise loss in Eq.~\ref{eq2:loss_hier}, using only available labels. 

%
\vspace{2pt}
\noindent\textbf{4) Implementation:}
We use H-ViT, a ViT-Small-based hierarchical classifier, as the backbone for evaluating both Text-Attr and \methodSSL. To evaluate its compatibility across architectures, we also apply Text-Attr to H-CAST~\citep{hcast2025park}, a state-of-the-art hierarchical model with comparable capacity. HRN~\citep{hrn2022chen} is evaluated with its original ResNet-50 backbone, which has over twice the parameters. All models are trained for 100 epochs, except for \dataINReal, which is trained for 200 due to its larger scale. Full architectural and training details are in the appendix~\ref{sec:suppl_implementation_details}.


\subsection{Benchmarking Results}\label{subsec:results_benchmark}
\noindent{\bf Result 1: Performance drop under free-grained learning.}  
The prior hierarchical SOTA, H-CAST, degrades sharply under mixed-granularity labels on both \dataCUB and \dataInat.  
Fig.~\ref{fig:performance_drop} shows that the full-path accuracy drops from 84.9\% to 45.1\% on \dataCUBReal and from 64.9\% to 25.6\% on \dataInatReal.  
This highlights the challenge of mixed-granularity labels and imbalanced supervision across the hierarchy, and the need for methods that handle them.

\noindent\textbf{Result 2: Performance on \dataINReal.}  
As shown in Table~\ref{tab:main_imagenet_real}, existing hierarchical methods degrade sharply under free-grained learning: HRN reaches only 37.8\% FPA, while H-CAST performs better at 57.6\% but still struggles with missing labels.  
\methodTextViT achieves 55.5\% without relying on H-CAST’s visual grouping, and integrating it into H-CAST further improves performance to 63.2\%, demonstrating the effectiveness of semantic-guided pseudo-attribute learning at scale.  
This is further supported by per-class gains in Appendix Sec.~\ref{sec:hcast_vs_hcastText}, particularly for classes with limited training samples.
\methodSSL improves over HRN by leveraging visual guidance but remains less effective than Text-Attr methods, whose strong performance benefits from the abundance and diversity of \dataINReal for reliable visual–semantic alignment.


\noindent\textbf{Result 3: Performance on \dataInatReal.}  
In Table~\ref{tab:main_imagenet_real}, on the large-scale \dataInatReal dataset, which contains many classes (10,000), conventional hierarchical methods perform poorly (17.0\% for HRN, 25.63\% for H-CAST). \methodSSL achieves the best performance (31.9\% FPA), highlighting the benefits of structural label propagation under limited per-class supervision. Text-Attr methods perform slightly lower (27.9–30.0\% FPA), likely due to restricted textual diversity in this fine-grained biological domain, yet still outperform conventional baselines. 

\noindent\textbf{Additional results and ablations.} 
We report additional results on \dataCUBReal (Sec.~\ref{sec:supple_cub_real}) and randomized variants with varying (limited) label availability (Sec.~\ref{sec:suppl_results_synthetic}). Across these settings, conventional hierarchical methods degrade under mixed-granularity supervision, while our approaches remain effective and robust. 
We further evaluate robustness under the original (unrefined) WordNet hierarchy, which exhibits irregular depth and inconsistent granularity (Sec.~\ref{supple:messy_hierarchy}). In addition, we conduct ablations on text encoders, Text-Attr features, training strategies, and architecture design (Sec.~\ref{sec:supple_ablation}), validating each component’s contribution.




\begin{figure}[!t]
\centering
 \includegraphics[width=0.7\linewidth]{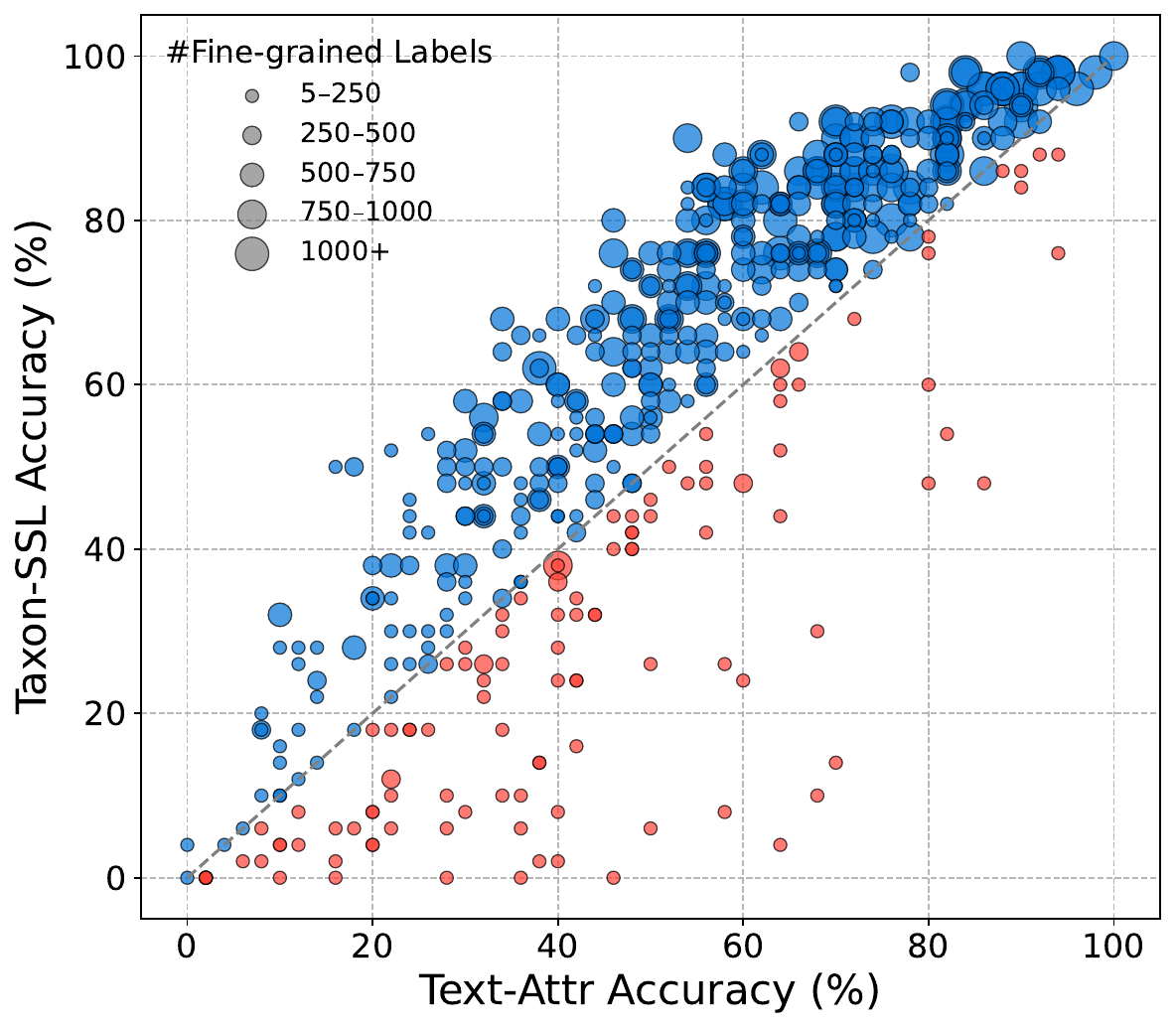}
 \vspace{-3mm}
 \caption{\textbf{Text-Attr is more effective under scarce fine-grained supervision, while \methodSSL performs better with more training data.} Each circle represents a class in \dataINReal, with Text-Attr fine-grained accuracy on the x-axis and \methodSSL accuracy on the y-axis. The diagonal marks equal performance: points below (\protect\redcircle) favor Text-Attr, and points above (\protect\bluecircle) favor \methodSSL. Circle size indicates the number of available training samples per class. Smaller circles tend to lie below the diagonal, showing the advantage of Text-Attr under limited data by leveraging textual guidance, whereas larger circles more often lie above it, indicating that \methodSSL benefits from richer supervision.}
\label{fig:clswise_acc_comparison}
\vspace{-7mm}
\end{figure}

\begin{figure}[t]
\small\centering
\newcommand{\squareimage}[1]{\includegraphics[width=0.2\linewidth, height=0.2\linewidth]{#1}}
\newcommand{\colspace}{\hspace{5pt}}
\newcommand{\colspaceclose}{\hspace{2pt}}

\newcommand{\tablerow}[1]{
\squareimage{figures/attention_imgs/#1.jpg} &
\squareimage{figures/attention_imgs/#1_CHMatchModule_basic.jpg} &
\squareimage{figures/attention_imgs/#1_CHMatchModule_subordinate.jpg} &
\squareimage{figures/attention_imgs/#1_CHMatchModule_finegrained.jpg} &
\squareimage{figures/attention_imgs/#1_HierVisionTransformer_basic.jpg} &
\squareimage{figures/attention_imgs/#1_HierVisionTransformer_subordinate.jpg} &
\squareimage{figures/attention_imgs/#1_HierVisionTransformer_finegrained.jpg}

}
\resizebox{1\linewidth}{!}{
\begin{tabular}
{@{}
c@{\colspace} 
c@{\colspaceclose}c@{\colspaceclose}c@{\colspace} 
c@{\colspaceclose}c@{\colspaceclose}c@{}
}
&
\multicolumn{3}{c}{\methodSSL} &
\multicolumn{3}{c}{Text-Attr} \\
\cmidrule(lr){2-4}\cmidrule(lr){5-7}
Image & 
\levelCoarse & \levelMid & \levelFG &
\levelCoarse & \levelMid & \levelFG \\
%
%
%

\tablerow{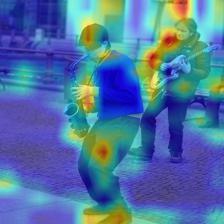}\\[-1pt]
&
clothing \redxmark & headwear \redxmark & saxophone \greencmark &
music inst. \greencmark & wind inst. \greencmark & saxophone \greencmark \\[1pt]
\tablerow{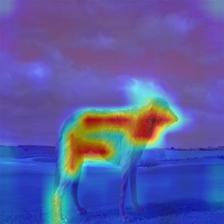}\\[-1pt]
&
dog \greencmark & hound dog \greencmark & megalith \redxmark &
dog \greencmark & hound dog \greencmark & {\footnotesize stand. poodle} \redxmark \\[1pt]
\end{tabular}
}\vspace{-2pt}
\caption{
\textbf{Text-Attr improves semantic focus under diverse large-scale data.}
(\textbf{Row 1}) In a multi-object image, \methodSSL assigns inconsistent labels (“\textit{clothing}" at the basic level, “\textit{saxophone}" at the fine-grained level), while \methodTextViT correctly predicts “\textit{musical instrument}” by focusing on the relevant object.  
(\textbf{Row 2}) When both fail at the fine-grained level, \methodSSL outputs an unrelated class (“\textit{megalith}”), whereas \methodTextViT chooses a semantically closer one (“\textit{poodle}”).  
This shows that text-derived attributes help the model attend to meaningful regions and maintain semantic plausibility, on a large-scale \dataINReal dataset with diverse categories and sparse labels. \textcolor{darkgreen}{Green}/\textcolor{darkred}{Red} denote correct/incorrect predictions.
}

\label{fig:saliency}
\vspace{-7mm}
\end{figure}

\subsection{Further Analysis}\label{subsec:analysis}
\noindent\textbf{How do methods behave with varying label availability?}
Text-attr excels with sparse labels, \methodSSL with moderate label availability.
We analyze class-wise performance under imbalanced fine-grained label availability on \dataINReal. To isolate effects, we compare \methodTextViT and \methodSSL with identical ViT-small backbones, excluding H-CAST modules. 
Fig.~\ref{fig:clswise_acc_comparison} plots per-class accuracy, where the x-axis shows Text-Attr performance and the y-axis shows \methodSSL performance; the diagonal indicates equal performance. 
\methodTextViT tends to outperform in label-scarce classes, appearing below the diagonal, by leveraging textual descriptions as additional supervision, while \methodSSL performs better for classes with more training samples, appearing above the diagonal by propagating consistency across missing levels. 
We provide additional class-wise breakdowns across different data scales in Sec.~\ref{sec:taxon_vs_textattr}, along with t-SNE~\citep{tsne2008} analysis in Sec.~\ref{sec:supple_tsne}.

\noindent\textbf{How does external semantic guidance help?}
External semantic guidance helps the model attend to semantically relevant features and improves hierarchical consistency.
To assess this effect, we compare saliency maps~\citep{chefer2021transformer} from \methodSSL and \methodTextViT (Fig.~\ref{fig:saliency}). 
In Row 1, with multiple objects, \methodSSL focuses on a human shoulder and misclassifies the image, violating the hierarchy, while Text-Attr attends to the instrument and predicts correctly.  
In Row 2, when both fail at the fine-grained level, \methodSSL predicts an unrelated class, whereas Text-Attr selects a visually similar dog by focusing on curly fur and body shape.  
These results show that text-derived semantic cues guide attention toward meaningful features across label granularities, while \methodSSL may drift to visually salient but semantically irrelevant regions under sparse or ambiguous supervision.

\subsection{Free-grained Inference}\label{subsec:free-inference}
\begin{figure}[t]
    \centering
    \includegraphics[width=0.9\linewidth]{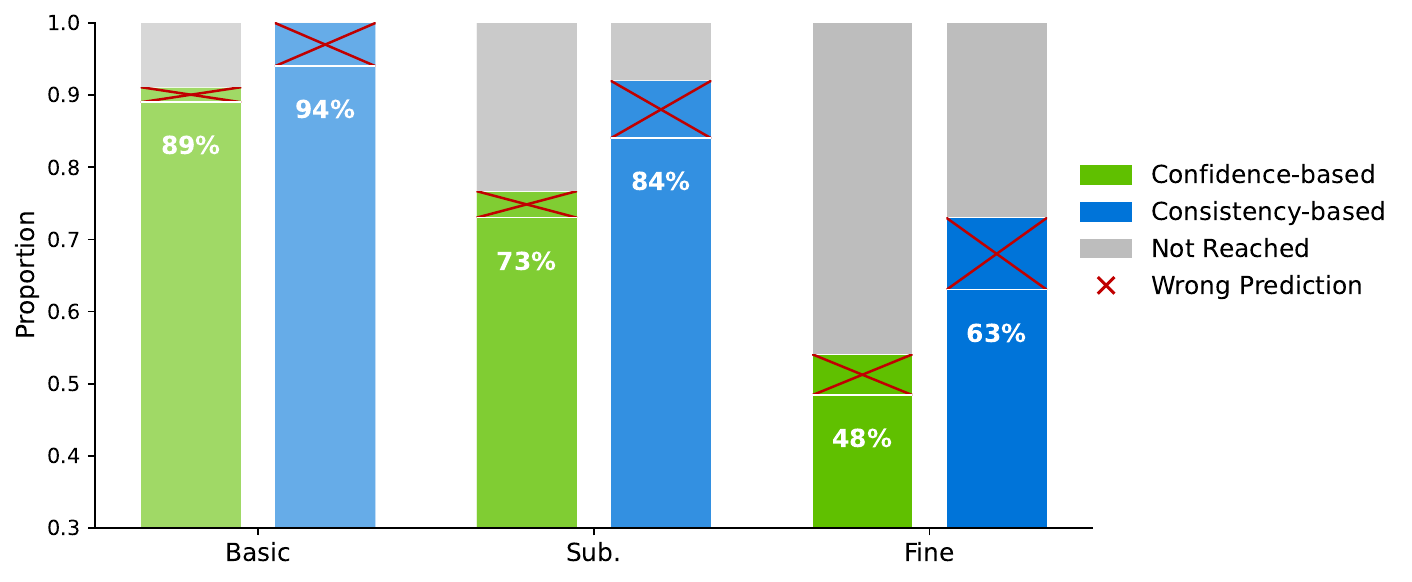}
    \vspace{-3mm}
    \caption{\textbf{Consistency-based stopping (blue) yields more reliable and deeper correct predictions than confidence-based stopping (green).} 
The figure compares confidence- and consistency-based stopping on \dataINReal dataset using \methodTextCAST.
Bars show the proportion of samples reaching each level (Basic → Subordinate → Fine). Gray (“Not Reached”) indicates early stopping at a coarser level, and red crosses mark incorrect predictions. 
Confidence-based rules often stop early, failing to reach deeper levels due to probability splitting among similar classes, whereas consistency-based stopping more often reaches deeper correct predictions.}
    \label{fig:confidence_vs_consistence}
\vspace{-3mm}
\end{figure}

Free-grained inference matters in practice, since a correct coarse label is often preferable to an incorrect fine-grained one. We compare confidence-based and consistency-based stopping. Confidence-based stopping halts when the softmax probability falls below $\tau\!=\!0.9$ (chosen from $[0.85, 0.99]$), but often stops prematurely because probability is split across similar sibling classes (Fig.~\ref{fig:confidence_vs_consistence}). 

\begin{figure}[t]
    \centering
\setlength{\tabcolsep}{0pt}
\begin{tabular}{@{}c@{}}
\includegraphics[width=1\linewidth]{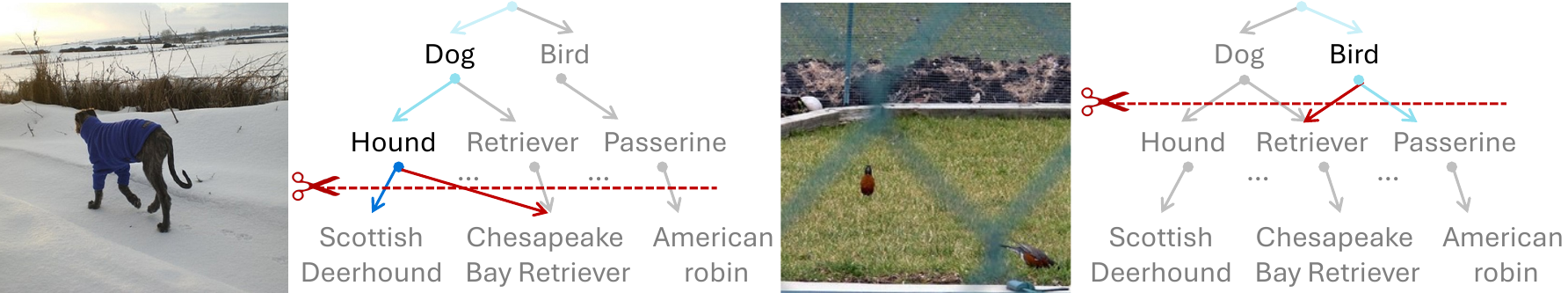}
\\
\includegraphics[width=0.95\linewidth]{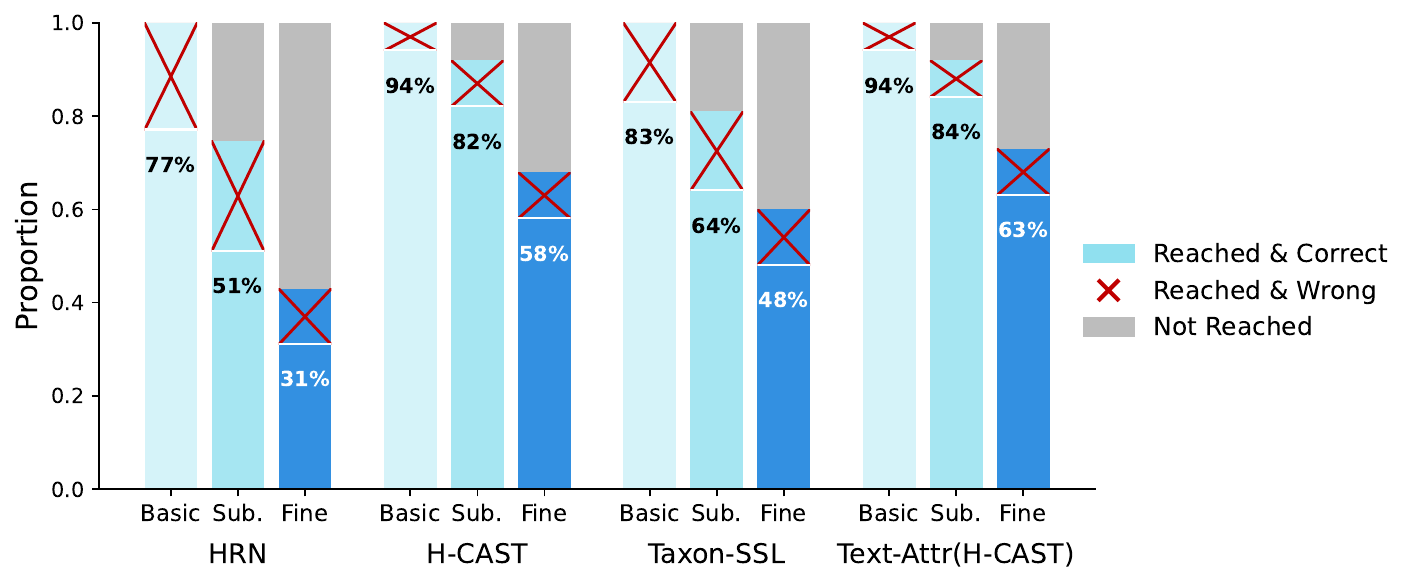}\\[-10pt]
\end{tabular}
\caption{\textbf{Consistency-based free-grained inference reliably stops at appropriate levels for low-confidence samples on \dataINReal.}
\textbf{Top:} The model stops at the appropriate level, subordinate (e.g., \textit{Hound}, left) or basic (e.g., \textit{Bird}, right), when deeper predictions become inconsistent, yielding more reliable outputs.
\textbf{Bottom:} Inference stops when finer-level predictions conflict with their coarser ancestors.
On \dataINReal, \methodTextCAST descends deeper while maintaining correctness, whereas HRN stops earlier and produces fewer fine-level predictions.}
\label{fig:free_grain_infer}
\vspace{-5mm}
\end{figure}

Consistency-based stopping halts only when taxonomy constraints are violated, requires no threshold tuning, and more reliably reaches deeper correct levels. It therefore yields more reliable and deeper correct predictions. Under this rule, \methodTextCAST produces the most taxonomy-consistent outputs (Fig.~\ref{fig:free_grain_infer}), reaching deeper correct levels while avoiding inconsistent fine-grained predictions. This suggests that stronger hierarchical consistency leads to more effective free-grained inference.

\vspace{-1mm}
\section{Conclusion}
\vspace{-1mm}
We introduce free-grained training and inference for hierarchical recognition, where models learn from labels of varying granularity while maintaining taxonomy consistency. We present diverse benchmarks and two methods that strongly outperform prior work, while highlighting the difficulty of free-grained learning and the need for more robust approaches.
\textbf{Limitations:}
Class- and level-wise imbalance are not addressed and are left for future work.
Additionally, better pruning based not just on CLIP could be explored to approximate intrinsic annotation difficulty.

\clearpage
\section*{Acknowledgements}
This project was supported, in part, by NSF 2215542, NSF 2313151, and Bosch gift funds to S. Yu at UC Berkeley and the University of Michigan, with additional compute support from NAIRR Pilot (CIS250430, CIS240431).


{
    \small
    \bibliographystyle{ieeenat_fullname}
    \bibliography{main}
}
\appendix

\onecolumn

\begin{center}{\bf \Large \papertitle}\end{center}
\begin{center}{\Large Supplementary Material}\end{center}

\hypersetup{linkcolor=black}

\etocdepthtag.toc{mtappendix}
\etocsettagdepth{mtchapter}{none}
\etocsettagdepth{mtappendix}{subsection}
\tableofcontents


\clearpage

\hypersetup{linkcolor=red}

\section{\dataIN-3L Dataset Construction}\label{sec:suppl_data_construction}

Our hierarchy design is inspired by cognitive studies~\citep{rosch1976basic}, which identify the \emph{basic level} (e.g., \textit{dog}) as the most natural and informative category for human recognition. Motivated by this, we construct a three-level hierarchy (\emph{basic}, \emph{subordinate}, and \emph{fine-grained}), starting from this most informative level and avoiding overly abstract categories (e.g., \textit{entity}, \textit{artifact}) that are less meaningful for visual recognition.

Importantly, the notion of the basic level is not fixed. As shown in~\cite{rosch1976basic}, category assignments can vary depending on context (e.g., ``\textit{fish}'' may appear at superordinate or basic depending on the experiment), suggesting that level boundaries are inherently flexible. We therefore adopt Rosch’s taxonomy as a guideline rather than a strict definition.

\noindent\textbf{1. Defining the Basic Level:}
We define the \emph{basic level} as categories that are both semantically meaningful and visually informative, ensuring comparisons at a consistent granularity. We primarily adopt categories aligned with Rosch’s superordinate level (e.g., \textit{bird}, \textit{musical instrument}), which better match the scale of ImageNet, while avoiding overly coarse concepts (e.g., \textit{entity}, \textit{animal}) and overly fine-grained ones. For example, we prefer \textit{dog}, \textit{bird}, and \textit{snake} over broader categories such as \textit{mammal} or \textit{reptile}.

Classes not covered by Rosch’s taxonomy are mapped to WordNet nodes at comparable semantic depths, and categories that are too coarse or too fine are adjusted to nearby levels. The resulting basic level is fixed globally, and all nodes above it are removed to enforce a uniform starting point. This avoids mismatched comparisons, such as between fine-grained label (e.g., \textit{teddy bear}) and coarse label (e.g., \textit{conveyance}), and ensures semantically coherent grouping across levels.

\noindent\textbf{2. Enforcing a Multi-level Hierarchy:}
We retain only categories that form meaningful three-level structures from \emph{basic} to \emph{subordinate} to \emph{fine-grained}. Categories that collapse into a single path are removed, including cases where nodes have only one child at each level or where the hierarchy terminates early. For example, if a basic-level category leads to only one subordinate class, which further has only one fine-grained class, the hierarchy provides no meaningful distinction across levels.
Similarly, we exclude shallow structures where the hierarchy does not extend to all three levels after defining the basic level. These pruning steps ensure that each retained category supports non-trivial branching and meaningful differentiation across levels.

\noindent\textbf{3. Selecting Categories for Diversity:}
When multiple valid candidates exist at a given level of the taxonomy, we select the category that leads to the richest set of descendant classes. This encourages greater intra-group diversity and supports more meaningful distinctions at finer levels.
For example, under \textit{bird}, both \textit{parrot} and \textit{cockatoo} are valid candidates. However, \textit{cockatoo} leads to only a single fine-grained class (e.g., \textit{sulphur-crested cockatoo}), whereas \textit{parrot} covers multiple diverse species (e.g., \textit{African grey}, \textit{sulphur-crested cockatoo}). We therefore select \textit{parrot} to ensure broader coverage and richer fine-grained classification.

\noindent\textbf{4. Handling Ambiguities and Naming:}
While WordNet provides a structured hierarchy, some categories are ambiguous or inconsistently defined. In such cases, we reorganize them using semantically coherent groupings. For example, instead of ill-defined categories such as \textit{Women’s Clothing}, we restructure them into functional groups (e.g., \textit{Underwear}) to improve clarity and consistency.

\noindent\textbf{5. Quality Control:}
We ensure taxonomy quality through a rule-based, human-in-the-loop process that verifies parent–child consistency and sibling-level coherence throughout construction. We further validate the structure using AI-assisted review (e.g., ChatGPT~\citep{achiam2023gpt4}) to identify potential inconsistencies or violations of intuitive categorization, followed by manual verification.

\begin{figure*}[h]
\centering

\begin{subfigure}[t]{0.48\linewidth}
    \centering
    \includegraphics[width=1\linewidth, keepaspectratio]{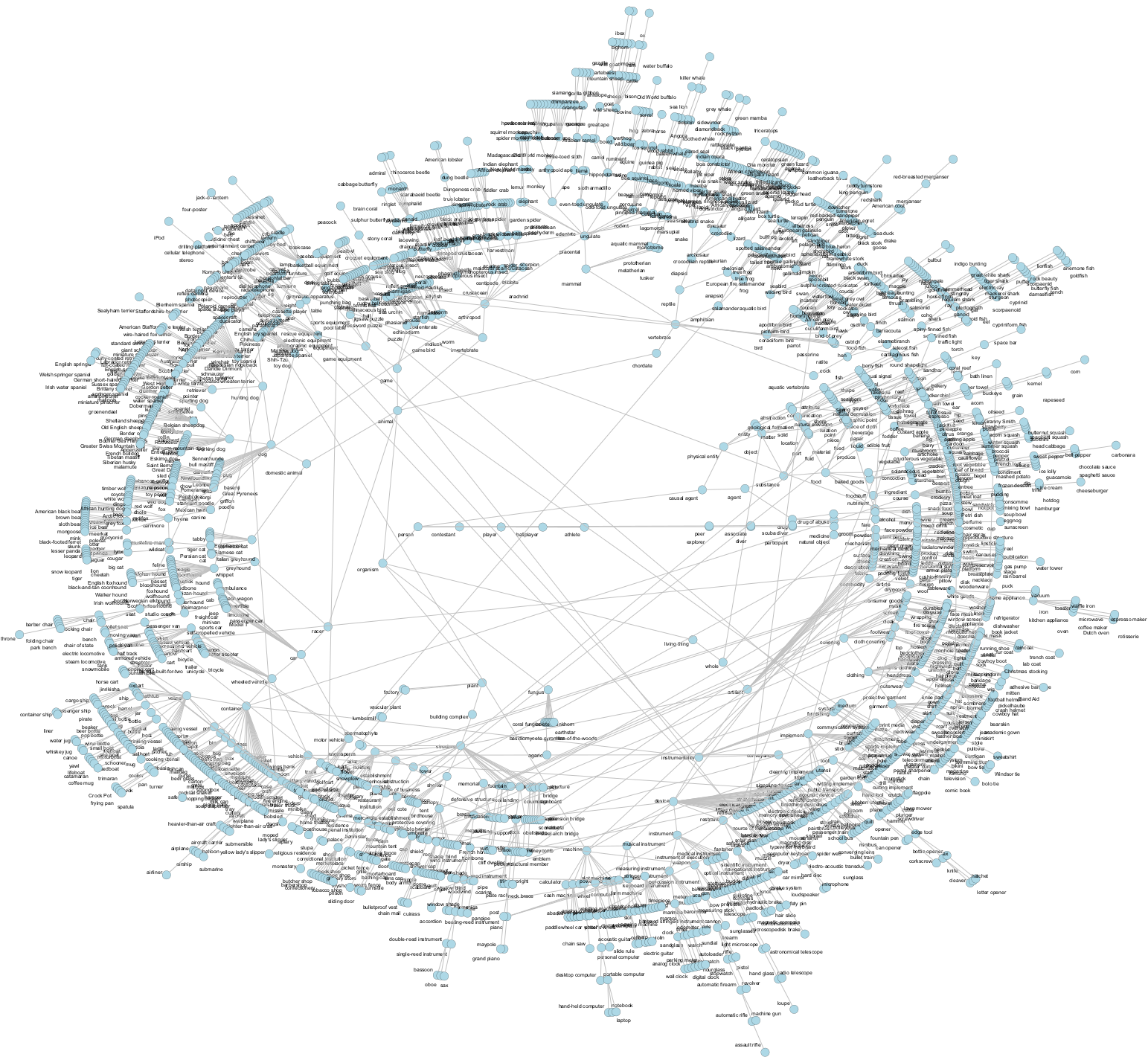}
    \subcaption{Original \dataIN's WordNet hierarchy}
    \label{fig:sub_a}
\end{subfigure}
\hfill
\begin{subfigure}[t]{0.48\linewidth}
    \centering
    \includegraphics[width=1\linewidth, keepaspectratio]{figures/3_level_circular_tree_final.png}
    \subcaption{Our 3-level hierarchy}
    \label{fig:sub_b}
\end{subfigure}

\caption{\textbf{We restructure the original WordNet hierarchy into a clean, consistent three-level hierarchy for hierarchical recognition.} }
\label{fig:main}
\end{figure*}

\section{Complete Hierarchy of \dataIN-3L}
\label{sec:suppl_complete_hierarchy}

{\small
\begin{longtable}{lp{.38\textwidth}p{.38\textwidth}}
 \toprule
 \textbf{Basic} & \textbf{Subordinate} & \textbf{Fine-Grained} \\

\midrule bird & passerine bird & brambling, indigo bunting, robin, jay, bulbul, water ouzel, house finch, chickadee, junco, magpie, goldfinch\\ 
\cmidrule{2-3} & parrot & macaw, sulphur-crested cockatoo, African grey, lorikeet\\ 
\cmidrule{2-3} & piciform bird & toucan, jacamar\\ 
\cmidrule{2-3} & seabird & king penguin, pelican, albatross\\ 
\cmidrule{2-3} & anseriform bird & drake, red-breasted merganser, black swan, goose\\ 
\cmidrule{2-3} & coraciiform bird & bee eater, hornbill\\ 
\cmidrule{2-3} & bird of prey & kite, great grey owl, vulture, bald eagle\\ 
\cmidrule{2-3} & gallinaceous bird & partridge, prairie chicken, ruffed grouse, peacock, quail, black grouse, ptarmigan\\ 
\cmidrule{2-3} & wading bird & flamingo, American coot, redshank, American egret, little blue heron, white stork, limpkin, spoonbill, red-backed sandpiper, dowitcher, crane, ruddy turnstone, bittern, oystercatcher, black stork, bustard\\ 
\midrule dog & spitz dog & malamute, Pomeranian, keeshond, Siberian husky, chow, Samoyed\\ 
\cmidrule{2-3} & pointer dog & vizsla, German short-haired pointer\\ 
\cmidrule{2-3} & spaniel dog & Brittany spaniel, clumber, English springer, Sussex spaniel, Irish water spaniel, Welsh springer spaniel, cocker spaniel\\ 
\cmidrule{2-3} & hound dog & basset, bloodhound, Irish wolfhound, Walker hound, redbone, English foxhound, Italian greyhound, Ibizan hound, bluetick, Scottish deerhound, borzoi, Norwegian elkhound, whippet, Weimaraner, Saluki, beagle, Afghan hound, black-and-tan coonhound, otterhound\\ 
\cmidrule{2-3} & terrier dog & Boston bull, silky terrier, Lakeland terrier, Yorkshire terrier, Tibetan terrier, American Staffordshire terrier, Irish terrier, Airedale, Norwich terrier, soft-coated wheaten terrier, wire-haired fox terrier, Staffordshire bullterrier, West Highland white terrier, Australian terrier, Dandie Dinmont, Kerry blue terrier, Lhasa, cairn, Sealyham terrier, Bedlington terrier, Scotch terrier, Border terrier, Norfolk terrier\\ 
\cmidrule{2-3} & corgi dog & Pembroke, Cardigan\\ 
\cmidrule{2-3} & poodle dog & miniature poodle, toy poodle, standard poodle\\ 
\cmidrule{2-3} & setter dog & Irish setter, Gordon setter, English setter\\ 
\cmidrule{2-3} & pinscher dog & Doberman, affenpinscher, miniature pinscher\\ 
\cmidrule{2-3} & shepherd dog & kelpie, briard, German shepherd, Old English sheepdog, Border collie, Bouvier des Flandres, collie, Rottweiler, komondor, malinois, groenendael, Shetland sheepdog\\ 
\cmidrule{2-3} & retriever dog & curly-coated retriever, Labrador retriever, Chesapeake Bay retriever, flat-coated retriever, golden retriever\\ 
\cmidrule{2-3} & schnauzer dog & standard schnauzer, miniature schnauzer, giant schnauzer\\ 
\cmidrule{2-3} & Sennenhunde dog & Bernese mountain dog, Greater Swiss Mountain dog, Appenzeller, EntleBucher\\ 
\cmidrule{2-3} & toy dog & toy terrier, Blenheim spaniel, Maltese dog, Shih-Tzu, papillon, Pekinese, Chihuahua, Japanese spaniel\\ 
\midrule fish & soft-finned fish & coho, tench, eel, goldfish\\ 
\cmidrule{2-3} & shark & tiger shark, great white shark, hammerhead\\ 
\cmidrule{2-3} & spiny-finned fish & anemone fish, puffer, lionfish, rock beauty\\ 
\cmidrule{2-3} & ray & stingray, electric ray\\ 
\cmidrule{2-3} & ganoid fish & sturgeon, gar\\ 
\midrule primate & ape & gibbon, siamang, orangutan, chimpanzee, gorilla\\ 
\cmidrule{2-3} & monkey & titi, langur, colobus, squirrel monkey, baboon, guenon, marmoset, macaque, spider monkey, patas, howler monkey, proboscis monkey, capuchin\\ 
\cmidrule{2-3} & lemur & Madagascar cat, indri\\ 
\midrule snake & colubrid snake & water snake, garter snake, green snake, night snake, hognose snake, ringneck snake, king snake, thunder snake, vine snake\\ 
\cmidrule{2-3} & elapid snake & sea snake, Indian cobra, green mamba\\ 
\cmidrule{2-3} & viper & diamondback, horned viper, sidewinder\\ 
\cmidrule{2-3} & boa snake & boa constrictor, rock python\\ 
\midrule salamander & newt & eft, common newt\\ 
\cmidrule{2-3} & ambystomid salamander & spotted salamander, axolotl\\ 
\midrule insect & beetle & dung beetle, weevil, leaf beetle, tiger beetle, ladybug, rhinoceros beetle, long-horned beetle, ground beetle\\ 
\cmidrule{2-3} & orthopterous insect & cricket, grasshopper\\ 
\cmidrule{2-3} & dictyopterous insect & cockroach, mantis\\ 
\cmidrule{2-3} & hymenopterous insect & bee, ant\\ 
\cmidrule{2-3} & butterflyinsect & cabbage butterfly, lycaenid, monarch, admiral, sulphur butterfly, ringlet\\ 
\cmidrule{2-3} & odonate insect & dragonfly, damselfly\\ 
\cmidrule{2-3} & homopterous insect & cicada, leafhopper\\ 
\midrule furniture & table & desk, dining table\\ 
\cmidrule{2-3} & baby bed & cradle, crib, bassinet\\ 
\cmidrule{2-3} & seat & rocking chair, barber chair, park bench, throne, folding chair, toilet seat, studio couch\\ 
\cmidrule{2-3} & lamp & table lamp\\ 
\cmidrule{2-3} & cabinet & china cabinet, medicine chest\\ 
\midrule musical instrument & wind instrument & ocarina, flute, panpipe, oboe, cornet, sax, harmonica, bassoon, French horn, trombone\\ 
\cmidrule{2-3} & stringed instrument & banjo, harp, violin, cello, acoustic guitar, electric guitar\\ 
\cmidrule{2-3} & percussion instrument & steel drum, gong, marimba, drum, chime, maraca\\ 
\cmidrule{2-3} & keyboard instrument & upright, grand piano, accordion, organ\\ 
\midrule scientific instrument & laboratory glassware & Petri dish\\ 
\cmidrule{2-3} & magnifier & loupe, radio telescope \\
\midrule sports equipment & ball & golf ball, baseball, basketball, croquet ball\\ 
\cmidrule{2-3} & gymnastic apparatus & parallel bars, balance beam, horizontal bar\\ 
\cmidrule{2-3} & weight & barbell, dumbbell\\
\midrule electronic equipment & telephone & dial telephone, pay-phone, cellular telephone\\ 
\cmidrule{2-3} & computer peripheral & printer, joystick, computer keyboard, mouse\\ 
\cmidrule{2-3} & audio device & tape player, cassette player, CD player, iPod\\ 
\cmidrule{2-3} & network device & modem\\ 
\cmidrule{2-3} & display device & monitor, screen\\ 
\midrule clothing & bottoms (skirts) & hoopskirt, sarong, miniskirt, overskirt\\ 
\cmidrule{2-3} & tops (sweaters) & sweatshirt, cardigan\\ 
\cmidrule{2-3} & outwear & trench coat, poncho, fur coat\\ 
\cmidrule{2-3} & swimwear & maillot, bikini, swimming trunks\\ 
\cmidrule{2-3} & face \& headwear & wig, sombrero, mortarboard, bonnet, mask, cowboy hat, bearskin\\ 
\cmidrule{2-3} & nightwear & pajama\\ 
\cmidrule{2-3} & protective wear & apron, knee pad, lab coat\\ 
\cmidrule{2-3} & dresses \& Gowns & gown\\ 
\cmidrule{2-3} & underwear & brassiere\\ 
\cmidrule{2-3} & footwear & sock, Christmas stocking\\ 
\cmidrule{2-3} & neckwear & bow tie, bolo tie, Windsor tie\\ 
\cmidrule{2-3} & traditional \& formal Wear & abaya, kimono, vestment, academic gown\\ 
\cmidrule{2-3} & wraps \& shawls & stole, feather boa\\ 
\midrule container & reservoir & water tower, rain barrel\\ 
\cmidrule{2-3} & bag & mailbag, plastic bag, backpack, purse\\ 
\cmidrule{2-3} & jug & water jug, whiskey jug\\ 
\cmidrule{2-3} & vessel & mortar, pitcher, tub, ladle, bucket, coffee mug\\ 
\cmidrule{2-3} & bottle & wine bottle, beer bottle, pop bottle, water bottle, pill bottle\\ 
\cmidrule{2-3} & basket & hamper, shopping basket\\ 
\cmidrule{2-3} & box & mailbox, carton, pencil box, chest, crate\\ 
\cmidrule{2-3} & glass & goblet, beer glass\\ 
\cmidrule{2-3} & shaker & saltshaker, cocktail shaker\\ 
\midrule cooking utensil & pan & frying pan, wok\\ 
\cmidrule{2-3} & cooker & Crock Pot\\ 
\cmidrule{2-3} & pot & teapot, caldron, coffeepot\\ 
\midrule structure & monument & brass, megalith, triumphal arch, obelisk, totem pole\\ 
\cmidrule{2-3} & religious building & church, mosque, boathouse, monastery, stupa\\ 
\cmidrule{2-3} & housing & yurt, cliff dwelling, mobile home\\ 
\cmidrule{2-3} & public building & planetarium, library\\ 
\cmidrule{2-3} & movable structure & sliding door, turnstile\\ 
\cmidrule{2-3} & supporting structure & plate rack, honeycomb, pedestal\\ 
\cmidrule{2-3} & fence & stone wall, picket fence, chainlink fence, worm fence\\ 
\cmidrule{2-3} & bridge & steel arch bridge, viaduct, suspension bridge\\ 
\cmidrule{2-3} & residential structure & palace\\ 
\cmidrule{2-3} & agricultural structure & greenhouse, barn, apiary\\ 
\cmidrule{2-3} & commercial stucture & toyshop, restaurant, cinema, confectionery, bookshop, grocery store, tobacco shop, bakery, butcher shop, barbershop, shoe shop\\ 
\cmidrule{2-3} & barrier & grille, bannister, breakwater, dam\\ 
\cmidrule{2-3} & institutional structure & prison\\ 
\midrule tool & hand tool & hammer, plunger, screwdriver\\ 
\cmidrule{2-3} & garden tool & lawn mower, shovel\\ 
\cmidrule{2-3} & cutter & cleaver, plane, letter opener, hatchet\\ 
\cmidrule{2-3} & power tool & chain saw\\ 
\cmidrule{2-3} & opener & corkscrew, can opener\\ 
\midrule craft & sailing vessel & trimaran, schooner, catamaran\\ 
\cmidrule{2-3} & boat & fireboat, canoe, yawl, gondola, speedboat, lifeboat\\ 
\cmidrule{2-3} & ship & wreck, pirate, container ship, liner\\ 
\cmidrule{2-3} & warship & aircraft carrier, submarine\\ 
\cmidrule{2-3} & aircraft & airliner, warplane, airship, balloon\\ 
\midrule vehicle & bicycle & bicycle-built-for-two, mountain bike \\
\cmidrule{2-3} & bus & minibus, school bus, trolleybus \\
\cmidrule{2-3} & car & ambulance, beach wagon, cab, convertible, jeep, limousine, Model T, racer, sports car \\
\cmidrule{2-3} & truck & fire engine, garbage truck, pickup, tow truck, trailer truck \\
\cmidrule{2-3} & van & minivan, moving van, police van \\
\cmidrule{2-3} & locomotive & electric locomotive, steam locomotive \\
\cmidrule{2-3} & military vehicle & half track\\
\cmidrule{2-3} & self-propelled vehicle & forklift, recreational vehicle, snowmobile, tank, tractor, golfcart, snowplow, go-kart, moped, streetcar, amphibious vehicle\\
\cmidrule{2-3} & handcart & barrow, shopping cart \\
\cmidrule{2-3} & sled & bobsled, dogsled \\
\cmidrule{2-3} & train & bullet train \\
\cmidrule{2-3} & wagon & horse cart, jinrikisha, oxcart \\
\cmidrule{2-3} & wheeled vehicle & freight car, motor scooter, tricycle, unicycle \\
\midrule weapon & gun & rifle, assault rifle, revolver, cannon\\ 
\cmidrule{2-3} & ranged weapon & missile, projectile\\

 \bottomrule
 \caption{\textbf{Complete hierarchy tree for our proposed \dataIN-3L dataset.}}
 \label{tab:suppl_complete_hierarchy}
\end{longtable}
}

\clearpage
\section{Experiments on Messy Hierarchy}\label{supple:messy_hierarchy}

\begin{figure}[h]
    \centering
    \includegraphics[width=1\linewidth]{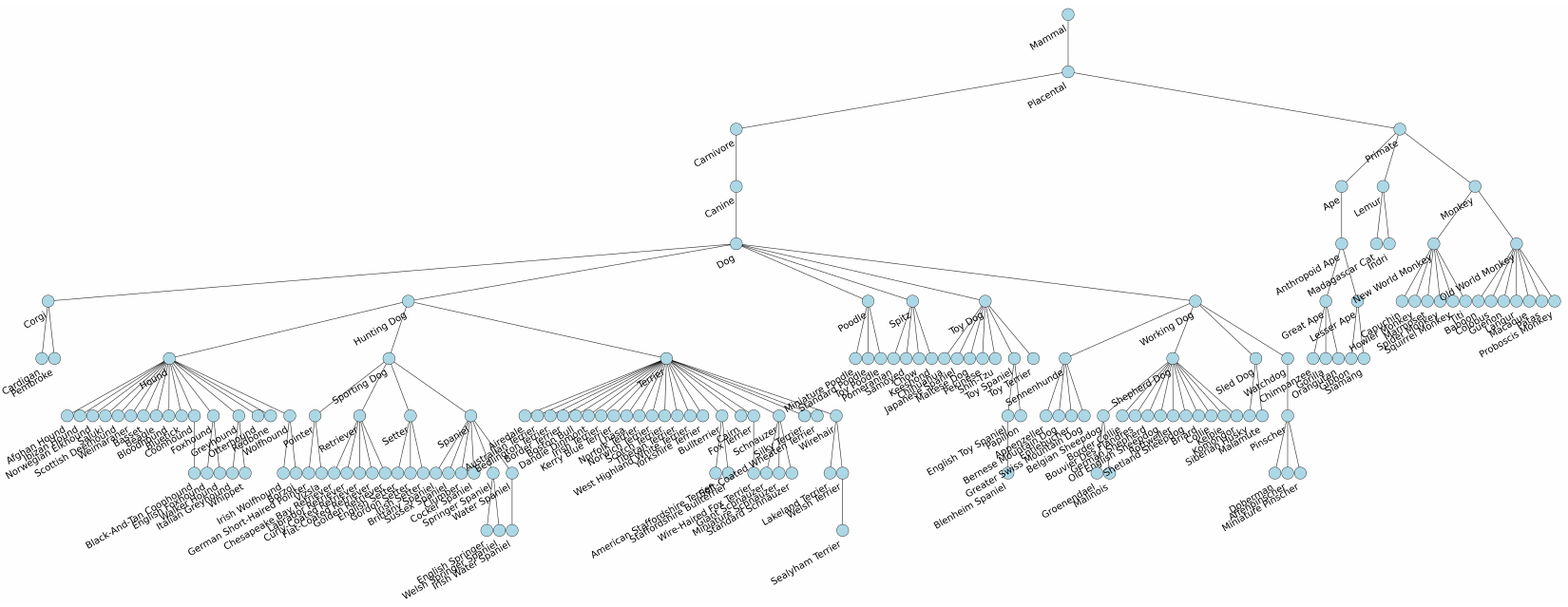}
    \caption{\textbf{Messy hierarchy of the ImageNet mammal subset.} The taxonomy exhibits irregular depths and inconsistent granularity across branches, making full-path prediction ill-posed. Despite these structural issues, our method consistently improves over the baseline, demonstrating robustness to noisy and misaligned hierarchies. Best viewed when zoomed in. }\label{fig:messy_hierarchy_mammal}
\end{figure}

\noindent\textbf{Setup.}
To evaluate robustness under the original WordNet structure, we construct a subset of ImageNet restricted to the \textit{mammal} subtree. All ancestor nodes above \textit{mammal} (e.g., \textit{vertebrate}, \textit{animal}, \textit{entity}) are removed, and only classes under \textit{mammal} are retained. This results in 120 fine-grained classes, with 152,548 training and 6,050 validation samples.
The resulting hierarchy remains highly irregular (Fig.~\ref{fig:messy_hierarchy_mammal}), with path lengths ranging from 5 to 10 levels. The distribution of chain lengths is skewed (e.g., most samples concentrate at depths 8–9, while deeper levels are sparse), leading to significant imbalance across levels.

\noindent\textbf{Limitations of the Raw WordNet Hierarchy for Full-Path Prediction.}
Our goal follows the standard objective of hierarchical recognition~\cite{chang2021fgn, hrn2022chen, hcast2025park}, where a model predicts the full path across all levels. However, the raw WordNet hierarchy is not well-suited for this setting due to its irregular structure: classes have varying depths, making consistent level alignment impossible.
To illustrate this, we construct a subset under the \textit{mammal} hierarchy and perform full-path prediction on the raw structure (Fig.~\ref{fig:messy_hierarchy_mammal}). We anchor all samples at the fine-grained level (Level 10) and propagate upward from \textit{mammal}, leaving intermediate levels empty when necessary. While this enables leaf-level comparison, it introduces semantic misalignment across higher levels. For example, \textit{hound} is compared with \textit{sporting dog} rather than \textit{spaniel}, and both \textit{spaniel} and \textit{English toy spaniel} appear at Level 8 despite their hierarchical inclusion.
These issues lead to (1) \textit{semantic inconsistency}, where nodes at the same level represent different levels of abstraction, and (2) \textit{structural sparsity}, where certain levels contain very few classes (e.g., Level 5–7), resulting in weak supervision. While such hierarchies can still support flat classification with hierarchical penalties, they are \textbf{not suitable for full-path prediction}, which requires consistent semantic alignment across levels. This motivates the need to restructure and align the hierarchy, as in our ImageNet-3L, so that each level corresponds to a coherent semantic granularity.

\noindent\textbf{Results and Analysis.}
Despite these challenges, \methodTextCAST consistently improves performance over \methodHCAST across most levels (Level 3–Level 10), demonstrating robustness to structural noise. Levels 1 and 2 are excluded, as each contains only one class, resulting in trivial (100\%) accuracy. The only exception is Level 9, where both methods perform poorly due to extreme sparsity and inconsistent supervision (some samples do not naturally have this level but are forced to predict it).
These results highlight that the gains from \textit{Text-Attr} are not tied to a well-structured hierarchy, but persist even in this ill-posed setting. At the same time, they motivate the need for a structured benchmark such as ImageNet-3L, which resolves these inconsistencies and provides a more meaningful evaluation of hierarchical recognition.

\begin{table}[h]
\caption{\textbf{Performance under the original WordNet hierarchy}, which exhibits irregular depth and highly imbalanced levels across classes, making it less suitable for hierarchical evaluation. 
Despite this, we train and evaluate on the given structure as a robustness test, where Text-Attr provides effective semantic supervision and improves performance.}
\label{table:messy_results}
\centering
\resizebox{0.8\linewidth}{!}{ 
\begin{tabular}{lcccccccc}
\hline
\begin{tabular}[c]{@{}c@{}}Depth\\ (\# class)\end{tabular}  & \begin{tabular}[c]{@{}c@{}}Level 10\\ (120)\end{tabular} & \begin{tabular}[c]{@{}c@{}}Level 9\\ (3)\end{tabular} & \begin{tabular}[c]{@{}c@{}}Level 8\\ (15)\end{tabular} & \begin{tabular}[c]{@{}c@{}}Level 7\\ (8)\end{tabular} & \begin{tabular}[c]{@{}c@{}}Level 6\\ (8)\end{tabular} & \begin{tabular}[c]{@{}c@{}}Level 5\\ (4)\end{tabular} & \begin{tabular}[c]{@{}c@{}}Level 4\\ (4)\end{tabular} & \begin{tabular}[c]{@{}c@{}}Level 3\\ (2)\end{tabular} \\ \hline
H-CAST           &   63.5   &     0.0    &   78.0    &   82.2   &   86.7  &   95.4   & 95.2  &   96.9  \\ \hline
\methodTextCAST &    \textbf{67.7}     &   0.0  &   \textbf{80.0}     &    \textbf{83.8}    &   \textbf{87.5}   & \textbf{96.6}   &   \textbf{96.0}      &   \textbf{98.0}   \\ \hline
\end{tabular}
}
\end{table}

\clearpage
\section{More Experimantal Results}
\subsection{Evaluaion on \dataCUBReal}\label{sec:supple_cub_real}

On the small-scale, single-domain dataset \dataCUBReal (Table~\ref{tab:suppl_cub_real}), \methodSSL achieves the best performance (63.96\% FPA), showing the advantage of structured label propagation when per-class samples are scarce.
Text-Attr methods perform moderately well (53.99–57.59\% FPA) but are less effective here, as the bird-only domain limits textual diversity and reduces the benefit of language-based supervision.
Still, they clearly outperform conventional hierarchical baselines (44.30\% for HRN, 45.10\% for H-CAST), underscoring the overall effectiveness of our approach.
Unlike the trend on large-scale, diverse datasets such as \dataINReal, where Text-Attr provides richer cues and stronger gains, these results confirm that there is no single recipe for free-grain learning: performance is tightly coupled with dataset characteristics, making the problem inherently challenging.

\begin{table}[h]
\caption{\textbf{\methodSSL shows strong effectiveness on the small-scale dataset \dataCUBReal, where label propagation provides reliable supervision.} Text-Attr methods are assumed to offer limited benefit due to the restricted textual diversity of this bird-only dataset.
}\label{tab:suppl_cub_real}
\centering
\resizebox{0.7\linewidth}{!}{
\begin{tabular}{l|ccccc}
\toprule
\dataCUBReal (13-38-200)              & FPA ($\uparrow$)& Species ($\uparrow$) & family ($\uparrow$) & Order ($\uparrow$)& TICE ($\downarrow$)\\ \midrule
\methodHRN~\citep{hrn2022chen}                    &  44.30 & 46.72   & 81.20 & 96.36     & 27.15   \\ 
\methodHCAST~\citep{hcast2025park}                &  45.10 & 47.52   & 87.78 & 97.50     & 25.89 \\  \midrule
\methodSSL             &  \textbf{63.96} & \textbf{65.50}   & \textbf{92.84} & \underline{98.40}     & \textbf{7.39}\\  
\methodSSL + Text-Attr &  \underline{63.05} & \underline{64.86}   & \underline{92.54} & 98.38     & \underline{7.61}\\  
\methodTextViT     &  {57.59} &	{59.10}&91.60	&{98.05}	&{10.72}\\ 
\methodTextCAST    &  53.99 & 55.58   & 91.72 & \textbf{98.41}     & 18.95\\ \bottomrule
\end{tabular}
}
\end{table}

\subsection{Evaluation under Varying and Severe Label Sparsity Conditions}\label{sec:suppl_results_synthetic}
To evaluate model performance under diverse free-grain conditions, we experiment with various label availability ratios by randomly removing multi-level labels, e.g., (100\%-60\%-30\%), (100\%-50\%-10\%), and (100\%-20\%-10\%), which represent the available proportions of basic, subordinate, and fine-grained labels, respectively. Each experiment is repeated with three different random seeds, and we report the average performance. The variance across runs was minor (0.1–1.8).

Consistent with our main results, these experiments (Table~\ref{tab:suppl_cub_air_6030} \&~\ref{tab:main_air_syn} \& ~\ref{tab:main_cub_syn}) also show that \textbf{there is \textit{no} single method that performs best across all settings}. Instead, the most effective method varies depending on the dataset and the specific ratio of available labels, highlighting the importance of adaptable free-grain learning strategies.

For consistency, we refer to the three levels in \dataCUBSyn (order-family-species) and \dataAirSyn (maker-family-model) as basic, subordinate, and fine-grained levels.
We summarize the key findings below:

\textbf{(1) Conventional hierarchical classification methods struggle under the free-grain setting, where label supervision is sparse and uneven across levels.}
For example, when labels are highly missing (e.g., only 10\% available at the fine-grained level), HRN~\citep{hrn2022chen} and H-CAST~\citep{hcast2025park} suffer more than a 50\% drop in accuracy across all levels compared to the fully labeled (100\%-100\%-100\%) setting on \dataCUBSyn (Fig. 6 \& Table~\ref{tab:main_cub_syn}).
This highlights the difficulty of the free-grain setting and the need for methods that can robustly handle incomplete supervision at multiple semantic levels.

\textbf{(2) The performance of different methods varies with the amount of available supervision per class: }
Text-Attr methods perform better when more labeled samples are available, while \methodSSL is more effective under extreme label sparsity.
For example, in Table~\ref{tab:suppl_cub_air_6030}, the average number of available fine-grained labels per class is approximately 9 for CUB-Rand and about 20 for Aircraft-Rand. Consistent with this difference, \methodSSL outperforms other methods on CUB-Rand, whereas \methodTextCAST performs best on Aircraft-Rand.
This trend persists across settings. In the most sparse setting, CUB-Rand (100-20-10, Table~\ref{tab:main_cub_syn}), where only about 3 fine-grained labels are available per class, \methodSSL shows a clear advantage.
We attribute this to how supervision is utilized. Text-Attr relies on available labels and indirect semantic guidance via text features. In contrast, \methodSSL actively leverages unlabeled data through pseudo-labeling and strong augmentations, making it more effective when labeled examples are extremely limited.

\textbf{(3) Sometimes, \methodSSL's high fine-grained accuracy comes at the cost of lower accuracy at higher levels in the taxonomy.}
For example, in Table~\ref{tab:main_air_syn}, \methodSSL achieves the highest fine-grained accuracy (65.01\%), but its subordinate and basic-level accuracies (85.53\% and 92.81\%) are lower than those of \methodTextCAST, which achieves 86.30\% and 94.17\%, respectively.
This highlights a key challenge in free-grain learning: improving accuracy across all levels simultaneously is non-trivial, and optimizing for fine-grained performance alone may degrade consistency at coarser levels.

\begin{table*}[h]
\caption{\textbf{No single method performs best across all conditions—performance depends strongly on the amount of available supervision per class.}
Text-Attr methods tend to perform better when more labeled samples are available, while \methodSSL is more effective under extreme label sparsity. For example, \methodSSL performs best on \dataCUBSyn with around 9 fine-grained labels per class, while \methodTextCAST performs best on \dataAirSyn with around 20, reflecting the impact of supervision density. 
These results highlight that method effectiveness is highly sensitive to label sparsity, emphasizing the need for adaptable approaches in free-grain learning.
}\label{tab:suppl_cub_air_6030}
\centering
\resizebox{0.95\linewidth}{!}{
\begin{tabular}{l 
                >{\centering\arraybackslash}p{0.9cm} 
                >{\centering\arraybackslash}p{0.9cm}
                >{\centering\arraybackslash}p{0.9cm}
                >{\centering\arraybackslash}p{0.9cm}
                >{\centering\arraybackslash}p{0.9cm} |
                >{\centering\arraybackslash}p{0.9cm}
                >{\centering\arraybackslash}p{0.9cm}
                >{\centering\arraybackslash}p{0.9cm}
                >{\centering\arraybackslash}p{0.9cm}
                >{\centering\arraybackslash}p{0.9cm}}
\toprule
Label Ratio& \multicolumn{5}{c|}{\dataCUBSyn(100\%-60\%-30\%)}              & \multicolumn{5}{c}{\dataAirSyn(100\%-60\%-30\%)}               \\ \cline{2-11} 
        & FPA($\uparrow$)   &  spec.($\uparrow$) & fam.($\uparrow$) & order($\uparrow$)  & {\small TICE($\downarrow$)}  & FPA($\uparrow$)   & {\small maker($\uparrow$)} & fam.($\uparrow$) & {\small model($\uparrow$)}  & {\small TICE($\downarrow$)}  \\ \midrule
\methodHRN\citep{hrn2022chen}     & 57.87 & 62.73   & 85.53 & 96.45 &13.77    &57.33 & 64.42   & 76.95 & 86.38     & 23.30 \\
\methodHCAST\citep{hcast2025park}  & 61.88 & 67.36  & 90.05 & 94.32     & 13.04 & 64.67 & 68.88   & 85.58 & 91.43     & 13.76 \\\midrule
\methodSSL & \underline{74.82} &\underline{76.92}   & \underline{93.38} & 98.33     & \underline{5.06} & \underline{70.33} & 72.22   & \underline{87.06} & \underline{93.50}     &\textbf{ 7.18} \\
\methodSSL + Text-Attr& \textbf{74.90}  &  \textbf{76.95}   & \textbf{93.41}  &  \underline{98.38 }    &  \textbf{4.91} &  69.89 &  \underline{72.24}   &  86.92 &   93.29    & \underline{7.77 } \\ 
\methodTextViT    & 67.89 & {72.48}   & 90.63 & 95.37     & 10.39 & 64.15 & 68.92   & {85.88} & 89.87     & 15.80 \\
\methodTextCAST    & {69.65} & 71.31   & {92.88} & \textbf{98.48}     & {8.35} & \textbf{71.43} & \textbf{73.56}   & \textbf{89.66} & \textbf{95.31}     & {9.71} \\ \bottomrule
\end{tabular}
}
\end{table*}
\begin{table*}[h]
\caption{\textbf{Maintaining accuracy across all hierarchy levels remains more challenging under sparse supervision.}
For example, in 100\%-50\%-10\% case, \methodSSL achieves the highest fine-grained accuracy (65.01\%), but its subordinate and basic-level accuracies (85.53\%, 92.81\%) are lower than those of \methodTextCAST (86.30\%, 94.17\%), which better preserves consistency across levels.
This result illustrates the inherent difficulty of improving accuracy across all levels simultaneously, as objectives at different levels can be conflicting.
}\label{tab:main_air_syn}
\centering
\resizebox{0.95\linewidth}{!}{
\begin{tabular}{l 
                >{\centering\arraybackslash}p{0.9cm} 
                >{\centering\arraybackslash}p{0.9cm}
                >{\centering\arraybackslash}p{0.9cm}
                >{\centering\arraybackslash}p{0.9cm}
                >{\centering\arraybackslash}p{0.9cm} |
                >{\centering\arraybackslash}p{0.9cm}
                >{\centering\arraybackslash}p{0.9cm}
                >{\centering\arraybackslash}p{0.9cm}
                >{\centering\arraybackslash}p{0.9cm}
                >{\centering\arraybackslash}p{0.9cm}}
\toprule
Label Ratio& \multicolumn{5}{c|}{\dataAirSyn(100\%-50\%-10\%)}              & \multicolumn{5}{c}{\dataAirSyn(100\%-20\%-10\%)}               \\ \cline{2-11} 
        & FPA($\uparrow$)   & {\small maker($\uparrow$)} & fam.($\uparrow$) & {\small model($\uparrow$)}  & {\small TICE($\downarrow$)}   & FPA($\uparrow$)   & {\small maker($\uparrow$)} & fam.($\uparrow$) & {\small model($\uparrow$)}  & {\small TICE($\downarrow$)}  \\ \midrule
\methodHRN~\citep{hrn2022chen}     & 40.35 & 47.85   & 70.76 & 85.68     & 37.56 & 32.06 & 46.73   & 55.43 & 85.58     & 48.43 \\
\methodHCAST~\citep{hcast2025park} & 47.57 & 51.93   & 78.31 & 87.11     & 28.42 & 40.33 & 45.44   & 67.28 & 84.12     & 35.61 \\ \midrule
\methodSSL & \underline{62.61} & \underline{65.01}   & {85.53} & \underline{92.81}     & \textbf{10.22} 
         & \textbf{58.73} & \textbf{61.10} & \underline{80.90} & \underline{92.24} & \textbf{11.77} \\ 
\methodSSL + Text-Attr& \textbf{62.95}  &  \textbf{65.49}   & \underline{86.01}  &  92.64     & \underline{10.25}  &  \underline{58.55} &  \underline{60.88}   & \textbf{80.97}  &  92.04     &  \underline{11.89} \\ 
\methodTextViT    & 47.83 & 52.25   & 81.13 & 87.82     & 30.57 
            & 38.73 & 43.89 & 66.13 & 84.81 & 38.69 \\
\methodTextCAST    & {53.31} & {55.32}   & \textbf{86.30} & \textbf{94.17}     & {24.43}
            & {48.85} & {51.37} & {77.11} & \textbf{93.01} & {27.25} \\
\bottomrule
\end{tabular}
}
\end{table*}

\begin{table*}[h]
\caption{\textbf{Taxon-SSL is more robust under extreme label sparsity.}
In CUB-Rand (100\%-20\%-10\%), where each class has only ~3 fine-grained and ~3 subordinate labels, Taxon-SSL achieves the best performance, while other methods struggle. HRN and H-CAST suffer over 50\% drop in fine-grained accuracy compared to the fully-supervised (100\%-100\%-100\%) setting. Text-Attr methods perform more robustly (10\%+ higher than HRN/H-CAST), but still struggle under sparse supervision.
We attribute this to how each method leverages supervision: Text-Attr depends on the provided labels and text-derived semantics, whereas \methodSSL can better exploit unlabeled data through pseudo-labeling and augmentations, leading to stronger performance when label sparsity is severe.
}\label{tab:main_cub_syn}
\centering
\resizebox{0.95\linewidth}{!}{
\begin{tabular}{l 
                >{\centering\arraybackslash}p{0.9cm} 
                >{\centering\arraybackslash}p{0.9cm}
                >{\centering\arraybackslash}p{0.9cm}
                >{\centering\arraybackslash}p{0.9cm}
                >{\centering\arraybackslash}p{0.9cm} |
                >{\centering\arraybackslash}p{0.9cm}
                >{\centering\arraybackslash}p{0.9cm}
                >{\centering\arraybackslash}p{0.9cm}
                >{\centering\arraybackslash}p{0.9cm}
                >{\centering\arraybackslash}p{0.9cm}}
\toprule
Label Ratio& \multicolumn{5}{c|}{\dataCUBSyn(100\%-50\%-10\%)}              & \multicolumn{5}{c}{\dataCUBSyn(100\%-20\%-10\%)}               \\ \cline{2-11} 
        & FPA($\uparrow$)   &  spec.($\uparrow$) & fam.($\uparrow$) & order($\uparrow$)  & {\small TICE($\downarrow$)}   & FPA($\uparrow$)   &  spec.($\uparrow$) & fam.($\uparrow$) & order($\uparrow$)  & {\small TICE($\downarrow$)}   \\ \midrule
\methodHRN~\citep{hrn2022chen}  & 40.23 & 43.70 &  82.75 & 95.94  & 22.34 
            & 33.53 & 41.18 &  72.56  & 95.79  & 30.50 \\
\methodHCAST~\citep{hcast2025park} & 39.03 & 43.41   & 85.74 & 93.23 & 24.60 
             & 32.97 & 38.66   & 76.89 & 92.50 & 29.43 \\ \midrule
\methodSSL & \underline{62.40} & \underline{64.14}  & \textbf{92.33} & \textbf{98.26} & \textbf{6.01}
           & \textbf{59.18} & \textbf{61.44} & \textbf{89.79} & \textbf{98.20} & \textbf{7.65} \\ 
\methodSSL + Text-Attr& \textbf{62.52}   &  \textbf{64.87}   &  87.94 &  94.45     & \underline{8.98}  &  \underline{57.98} &  \underline{60.59}   & \underline{89.42}  &  \underline{98.12}     & \underline{8.39}  \\ 
\methodTextViT  & {47.42} & {50.74} & 88.22 & 94.67 & {18.09}
                & {42.46} & {46.99} & 80.92 & 94.43 & {20.27} \\
\methodTextCAST & 44.63 & 45.89 & \underline{91.06} & \underline{98.19} & 22.72
                & 40.41 & 42.76 & {84.24} & {97.97} & 24.05 \\ 
\bottomrule
\end{tabular}
}
\end{table*}

\clearpage

\subsection{Effect of Training Data Scale on Method Preference}\label{sec:taxon_vs_textattr}

\begin{figure*}[h]
    \centering
    \includegraphics[width=\textwidth]{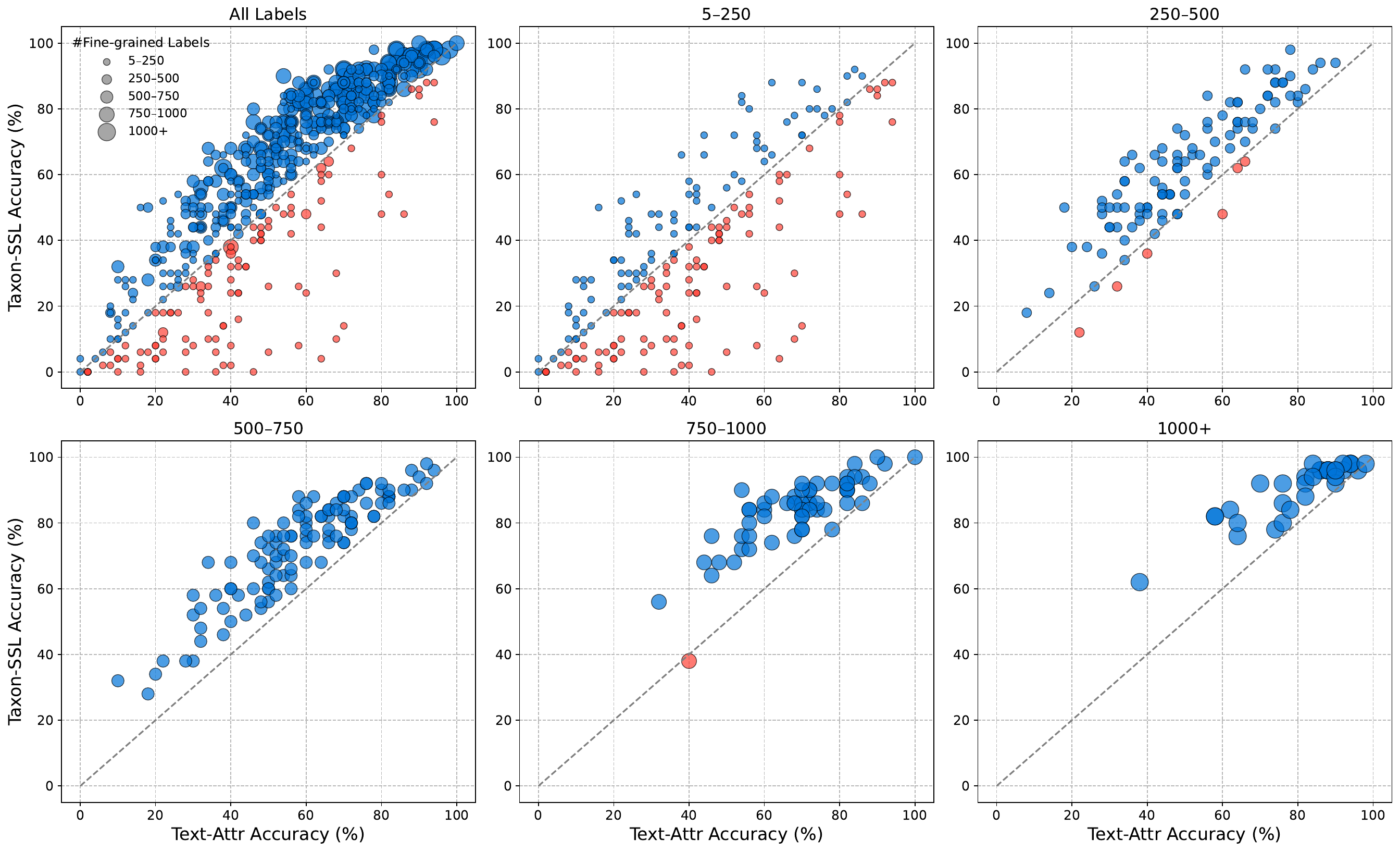}
    \caption{
    \textbf{\methodTextFull is especially beneficial for classes with very limited training data, whereas Taxon-SSL becomes increasingly favorable as class frequency grows.}
    The first panel shows the overall class-wise comparison between \methodTextViT and Taxon-SSL, and the other five panels present the same comparison separately for different ranges of fine-grained training sample counts per class (\mbox{5--250}, \mbox{250--500}, \mbox{500--750}, \mbox{750--1000}, and \mbox{1000+}). Each dot corresponds to one class. Blue dots denote classes where Taxon-SSL achieves higher accuracy, and red dots denote classes where \methodTextViT{} achieves higher accuracy; the dashed diagonal indicates equal accuracy.
    The plots show a clear regime-dependent trend: in the lowest-count regime (\mbox{5--250}), \methodTextViT{} frequently yields larger gains, while in higher-count regimes Taxon-SSL more often performs better. This suggests that text-guided pseudo-attributes are particularly helpful when fine-grained labels are scarce, whereas pseudo-label-based semi-supervised learning is more effective when enough training examples are available.
    }
    \label{fig:classwise_compare_by_count}
\end{figure*}

We analyze how the relative performance of \methodTextViT{} and Taxon-SSL varies with the number of training samples per class. To ensure a fair comparison, both methods share the same ViT backbone. Figure~\ref{fig:classwise_compare_by_count} presents a class-wise comparison, along with breakdowns across different ranges of fine-grained sample counts.

A clear trend emerges depending on the amount of training data. When training data per class is limited (5--250 samples), \methodTextViT{} more frequently outperforms Taxon-SSL, indicating that text-guided pseudo-attributes provide stronger supervision when labeled data are scarce. In contrast, as the number of training samples increases, the advantage gradually shifts toward Taxon-SSL, suggesting that pseudo-label-based semi-supervised learning becomes more effective with sufficient visual data.

These results highlight that the two approaches are complementary: \methodTextViT{} is particularly beneficial for classes with limited training data, while Taxon-SSL is better suited for classes with more abundant samples.

\clearpage
\subsection{Performance Gains under Varying Label Availability}\label{sec:hcast_vs_hcastText}

\begin{figure}[h]
    \centering
    \includegraphics[width=1\linewidth]{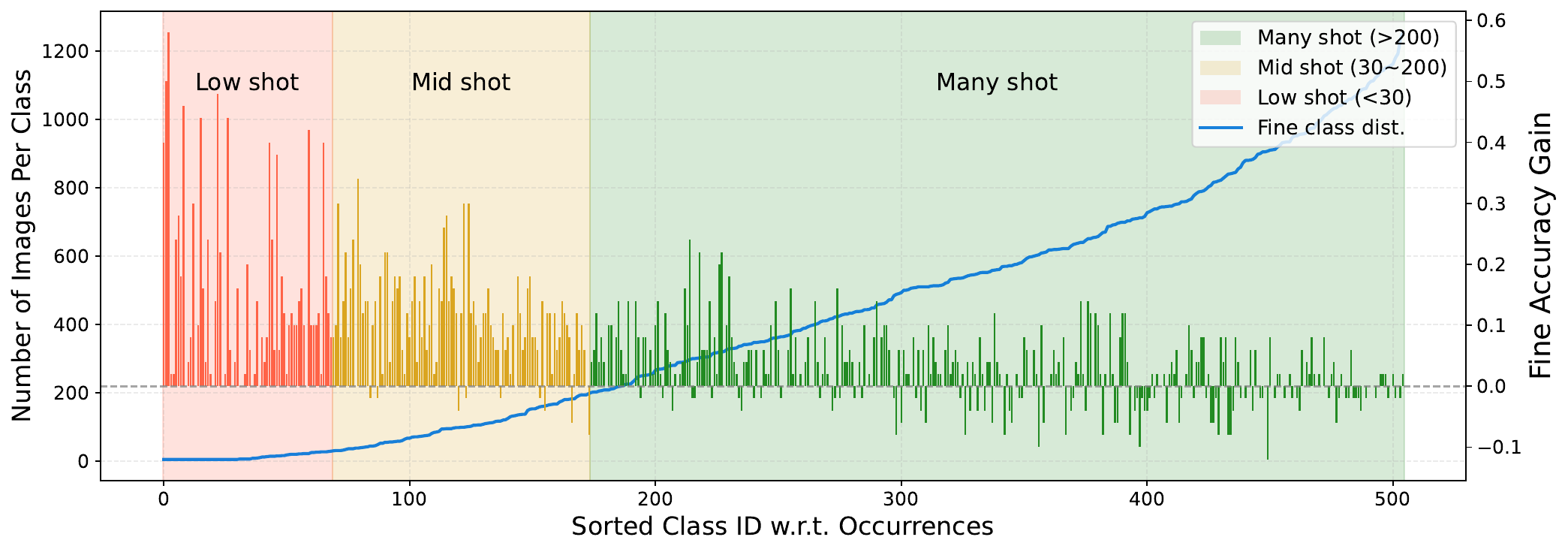}
    \caption{\textbf{\methodTextCAST achieves the largest accuracy gains over \methodHCAST in low- and mid-shot regimes on \dataINReal.}
Per-class fine-grained accuracy gains are shown, with classes sorted by the number of training samples. Background colors indicate low-shot ($<$30), mid-shot (30–200), and many-shot ($>$200) regimes. The blue curve shows the number of training samples per class. Gains are most significant in low- and mid-shot classes, while improvements in many-shot classes are smaller and but consistent. This demonstrates that text-guided features are particularly beneficial under limited supervision.}\label{fig:hcast_vs_hcastText}
\end{figure}

{Our method \methodTextCAST achieves the largest performance gains over \methodHCAST in low- and mid-shot regimes, demonstrating the effectiveness of text-guided features under limited supervision.}
Fig.~\ref{fig:hcast_vs_hcastText} shows the per-class fine-grained accuracy gain of \methodTextCAST over \methodHCAST on \dataINReal, with classes sorted by the number of training samples. Classes are grouped into three regimes: low-shot ($<$30 samples), mid-shot (30–200), and many-shot ($>$200). 

As shown in Fig.~\ref{fig:hcast_vs_hcastText}, the gains are most pronounced in low- and mid-shot classes, where limited visual data makes learning discriminative features challenging. In these regimes, incorporating text-derived semantic cues provides complementary supervision, leading to substantial improvements. In contrast, gains in the many-shot regime remain consistent but smaller, as sufficient visual data is already available.

These results highlight that text-guided features are particularly beneficial under data scarcity, enabling more robust fine-grained recognition when training samples are limited.

\clearpage
\section{t-SNE Visualization}\label{sec:supple_tsne}
We visualize \dataINReal embeddings of \methodTextCAST and \methodSSL using t-SNE~\citep{tsne2008} to assess whether the learned representations capture semantic and hierarchical structure. Each point denotes an image embedding, colored by its basic-level class (20 categories), with brightness variations indicating fine-grained subclasses (505 total).  

Both \methodTextCAST and \methodSSL produce well-separated clusters consistent with the basic-level taxonomy, showing that coarse groupings are reliably captured. The key difference lies within coarse categories: \textbf{\methodTextCAST reveals more distinct fine-grained subclusters} (e.g., breeds within \textit{dog}, species within \textit{bird}), whereas \textbf{\methodSSL yields tighter coarse clusters with less apparent fine-level separation}.  

This contrast reflects their supervision signals. Text-Attr leverages diverse textual cues (attributes, parts, appearance terms), which promote discriminative, attribute-aligned features and sharpen within-class distinctions. Taxon-SSL, by propagating labels along the taxonomy and enforcing consistency under mixed-granularity supervision, regularizes embeddings within each coarse class and reduces intra-class variance—emphasizing coarse alignment over fine-level separability.

\begin{figure*}[h]
\centering
\begin{subfigure}{0.45\linewidth}
    \includegraphics[width=\linewidth]{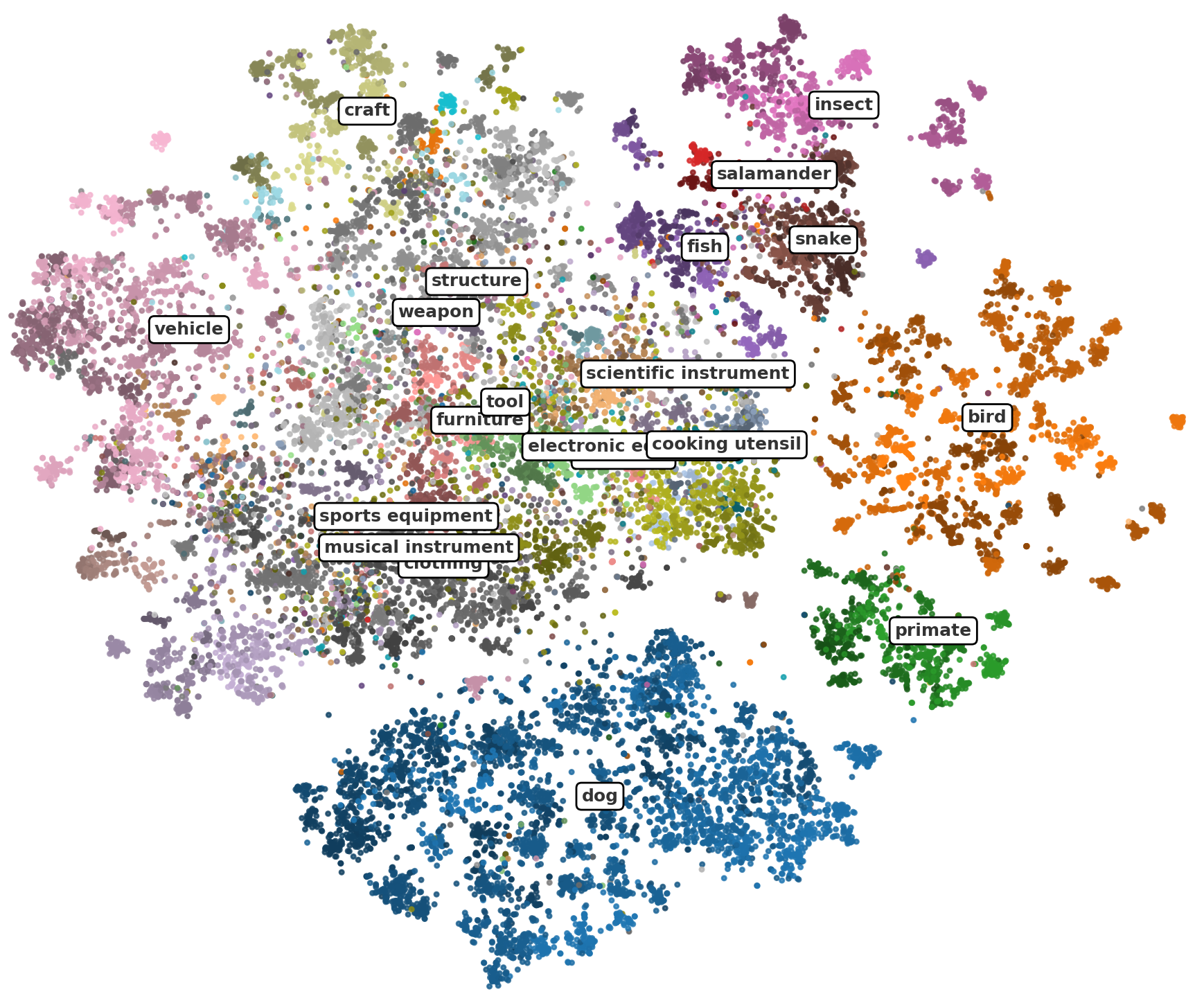}
    \caption{\methodTextCAST}
\end{subfigure}
\begin{subfigure}{0.5\linewidth}
    \includegraphics[width=\linewidth]{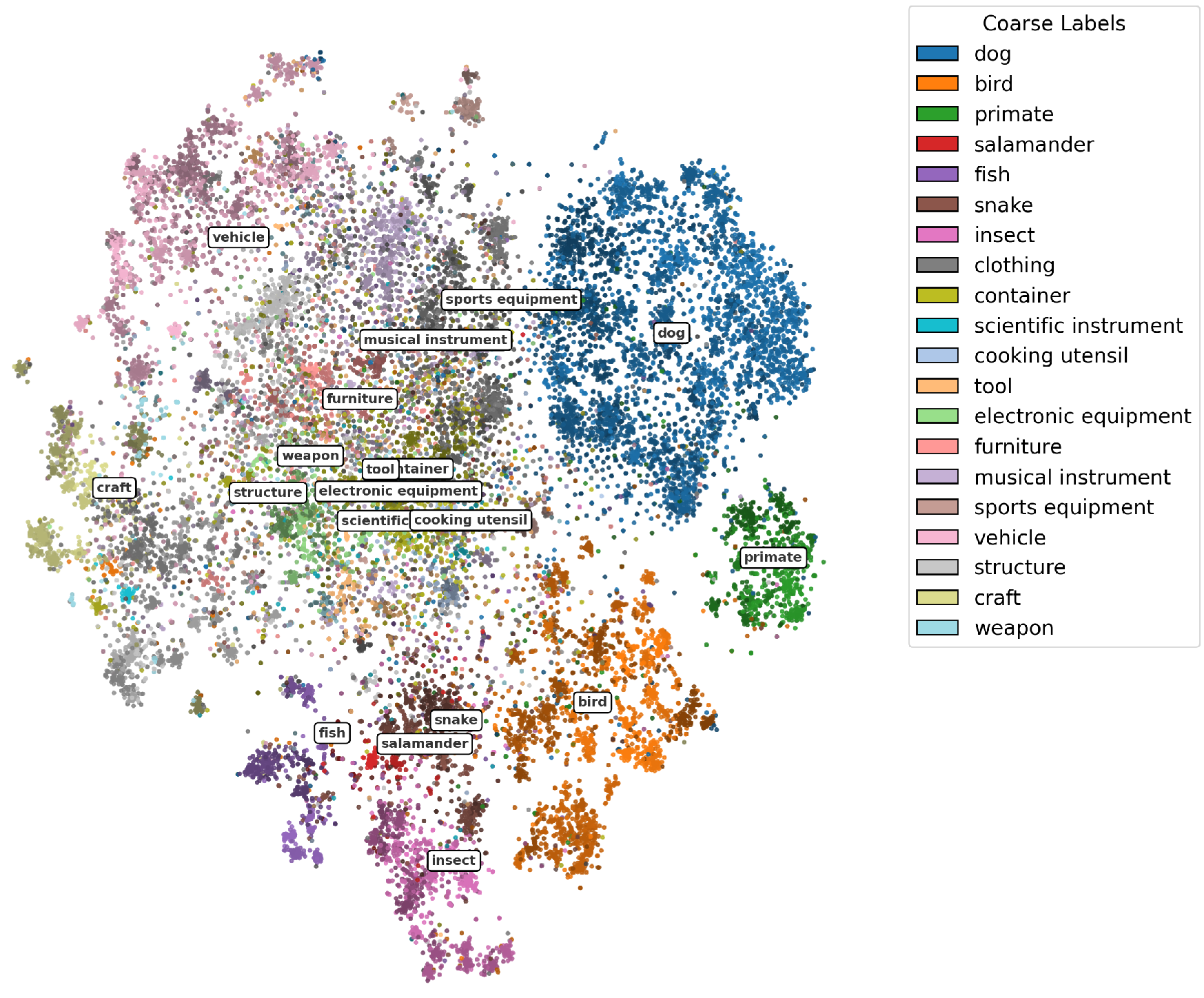}
    \caption{\methodSSL}
\end{subfigure}

\caption{\textbf{t-sne Visualization on \dataINReal.} Both methods separate coarse-level taxonomy well, but \methodTextCAST yields clearer fine-grained subclusters (e.g., distinct groups within \textit{dog} and \textit{bird}) with more compact grouping, whereas \methodSSL shows some overlap of embeddings near cluster boundaries. This is likely due to \dataINReal’s diverse large-scale categories, where text supervision provides rich attribute cues that sharpen fine-level distinctions.}
\label{fig:supple_tsne}
\end{figure*}

\clearpage
\section{Ablation Study}
\label{sec:supple_ablation}

\subsection{Importance of \methodTextFull}
\begin{figure}[h]
\centering
 \includegraphics[width=0.5\linewidth]{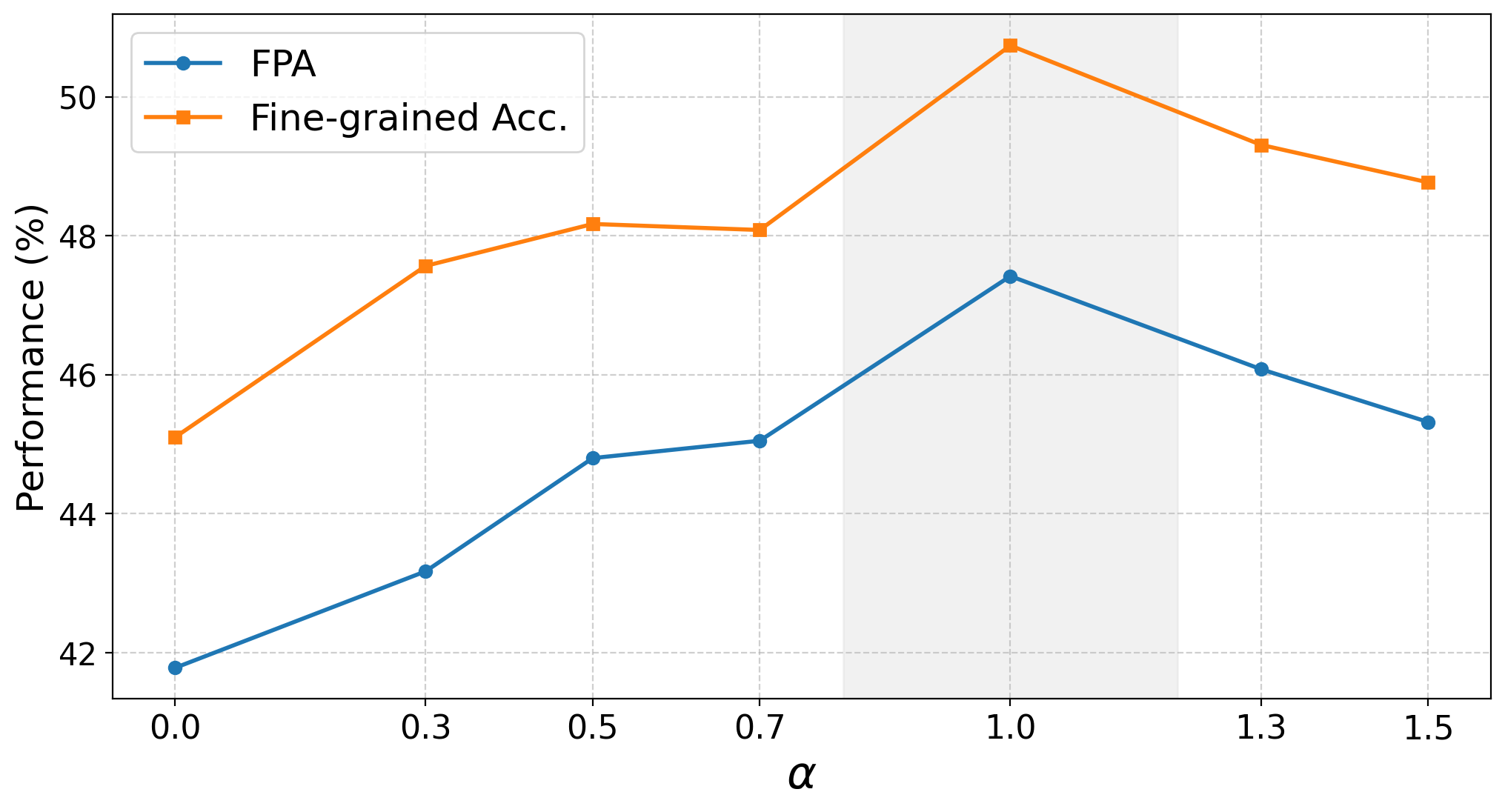} 
\caption{
\textbf{Tuning $\alpha$ balances accuracy and taxonomic consistency.} At $\alpha=1$ (optimal), \methodTextViT achieves peak fine-grained accuracy (blue) while maintaining hierarchical consistency (orange). Ablating $\mathcal{L}_{\text{text}}$ ($\alpha=0$) causes a 5\% accuracy drop and increased inconsistency, as class embeddings lose text-guided attribute alignment. Higher $\alpha >1.0$ over-regularizes features, marginally degrading both metrics. This trade-off underscores the need to weight text supervision to resolve sparse annotations without distorting the hierarchy.}
\label{fig:impact_weight}
\end{figure}

\methodTextFull jointly optimizes hierarchical label supervision ($\mathcal{L}_{\text{hier}}$) and text-guided pseudo attributes ($\mathcal{L}_{\text{text}}$) to learn semantically rich features: 
$\mathcal{L} = \mathcal{L}_{\text{hier}} + \alpha \mathcal{L}_{\text{text}}$
Fig.~\ref{fig:impact_weight} quantifies $\mathcal{L}_{\text{text}}$'s impact by varying its weight $\alpha$ on \dataCUBSyn. Ablating $\mathcal{L}_{\text{text}}$ ($\alpha=0$) causes a 5\% absolute decline in both fine-grained accuracy and FPA compared to the optimal configuration ($\alpha=0$). This gap underscores two key roles of text guidance: (1) it injects complementary visual semantics absent in class labels alone, and (2) it enforces attribute consistency across hierarchy levels. The performance recovery at ($\alpha=1$) confirms that textual pseudo-attributes mitigate annotation sparsity while preserving taxonomic coherence.

\subsection{Combining Text-Attr and \methodSSL}

We compare different training schedules for combining Text-Attr and \methodSSL on \dataCUBReal. 
In the \textbf{joint setting}, both objectives are optimized simultaneously for 100 epochs. 
In the \textbf{two-stage setting}, we first train with one objective for 50 epochs and then add the other for the remaining 50 epochs, considering both orders: (1) \methodSSL $\rightarrow$ Text-Attr, and (2) Text-Attr $\rightarrow$ \methodSSL.

Table~\ref{tab:suppl_ab_combining_training} show that starting with Text-Attr and then adding \methodSSL yields slightly higher full-path accuracy, likely because textual supervision promotes diverse feature learning before label propagation. 
In contrast, beginning with \methodSSL provides no advantage, and both two-stage variants perform similarly to joint training overall. 
Interestingly, joint training achieves higher consistency as measured by TICE. 
Given its simplicity and competitive performance, we adopt the joint strategy as our default.

\begin{table}[h]
\caption{\textbf{Comparison of joint vs. two-stage training schedules for Text-Attr and \methodSSL on \dataCUBReal.} While two-stage training (Text-Attr $\rightarrow$ \methodSSL) yields slightly higher accuracy, joint learning is simpler and provides better consistency (TICE).
}\label{tab:suppl_ab_combining_training}
\centering
\resizebox{0.8\linewidth}{!}{
\begin{tabular}{l|ccccc}
\toprule
\dataCUBReal (13-38-200)              & FPA ($\uparrow$)& Species ($\uparrow$) & family ($\uparrow$) & Order ($\uparrow$)& TICE ($\downarrow$)\\ \midrule
\methodSSL + Text-Attr (100 epochs)                           & \underline{63.04} & \underline{64.86}   & \underline{92.54}  & \textbf{98.37} & \textbf{7.61} \\ 
\methodSSL (50 epochs) $\rightarrow$ +Text-Attr (50 epochs)   & 62.84 & 64.42   & 92.47  & 98.20  & 8.19 \\ 
Text-Attr (50 epochs) $\rightarrow$ +\methodSSL (50 epochs)   & \textbf{63.63} & \textbf{65.34}   & \textbf{92.56}  & \underline{98.27} & \underline{8.06} \\ 
\bottomrule
\end{tabular}
}
\end{table}

\clearpage
\subsection{Ablation on the Text Encoder in Text-Attr}\label{supple:ab_textencoder}
Text-Attr relies on a text encoder to embed visual descriptions. To assess whether performance depends on the choice of text encoder, we replace CLIP with alternative encoders, including SigLIP~\cite{zhai2023siglip} (\textit{siglip-base-patch16-224}) and E5~\cite{wang2022e5} (\textit{e5-base-v2}). All models use base configurations with a 768-dimensional embedding, and are integrated into our framework with minimal modification.
As shown in Table~\ref{table:ablation_textencoder}, all encoders achieve comparable performance, with only minor differences across metrics. The text-only model (E5) performs slightly worse than CLIP and SigLIP, but the overall trend remains consistent.
These results suggest that the gains of Text-Attr are not tied to a specific text encoder, but rather stem from the proposed training formulation for learning visual attributes.

\begin{table}[h]
\caption{\textbf{Ablation on the text encoder used in Text-Attr on \dataINReal}. Replacing CLIP with SigLIP and E5 results in comparable performance, with a slight drop for the text-only encoder (E5), indicating that gains mainly come from the proposed training formulation rather than the choice of encoder.}\label{table:ablation_textencoder}
\centering
\resizebox{0.5\linewidth}{!}{
\begin{tabular}{lccccc}
\toprule
Text encoder  & FPA ($\uparrow$)  & fine ($\uparrow$) & sub. ($\uparrow$) & basic ($\uparrow$) & TICE ($\downarrow$)\\ \midrule
CLIP   & \textbf{63.2} & \textbf{64.9}    & \textbf{84.5}   & {93.6} & \textbf{18.6} \\
SigLIP & {62.9} & \textbf{64.9}    & {84.3}   & \textbf{93.7} & 19.5\\
E5     & 61.8 & 63.9    & 84.1   & 93.4 & 20.6 \\ \bottomrule
\end{tabular}
}
\end{table}

\subsection{Ablation on Hierarchical Supervision in ViT}

We further examine the architectural design choice of where to inject hierarchical supervision in the Vision Transformer (ViT) in Table~\ref{tab:suppl_ab_vit_arch}.
On \dataCUBReal, we map the three taxonomy levels (Order–Family–Species) to different layers and compare multiple configurations: (6th, 9th, 12th), (8th, 10th, 12th), and (10th, 11th, 12th). 

Among these, supervision at the 8th, 10th, and 12th layers yields the best performance. 
We interpret this as a balance between early and late representation learning:  
assigning hierarchy too early (e.g., 6–9–12) forces the model to align coarse categories before sufficient visual features are developed, while placing all supervision too late (e.g., 10–11–12) limits the model’s capacity to gradually refine class granularity. 
The 8–10–12 configuration provides an appropriate middle ground, where lower-level categories benefit from moderately abstract features, and finer distinctions are introduced after the backbone has matured.

\begin{table}[h]
\caption{\textbf{Performance comparison of different layer assignments for hierarchical supervision in ViT on \dataCUBReal.} The 8th–10th–12th configuration achieves the best results, balancing early and late feature abstraction.
}\label{tab:suppl_ab_vit_arch}
\centering
\resizebox{0.65\linewidth}{!}{
\begin{tabular}{l|ccccc}
\toprule
\dataCUBReal (13-38-200)   & FPA ($\uparrow$)& Species ($\uparrow$) & family ($\uparrow$) & Order ($\uparrow$)& TICE ($\downarrow$)\\ \midrule
6-9-12th layer                           & 54.80&	58.16&	88.97	&95.01	&16.79 \\ 
8-10-12th layer    & \textbf{57.59} &	\textbf{59.10}&\textbf{91.60}	&\textbf{98.05}	&\textbf{10.72}\\ 
10-11-12th layer   & 56.40&	58.56&	90.80	&97.08	&13.48 \\ 
\bottomrule
\end{tabular}
}
\end{table}

\clearpage
\subsection{Ablation on Captioning Strategy and Caption Cost}
Text-Attr relies on VLM-generated descriptions to capture visual attributes. To evaluate the benefit and cost of this step, we compare our approach with a simpler alternative that uses class-level text prompts (e.g., “a photo of [deepest available label]”) instead of image-level descriptions.
As shown in Table~\ref{table:ablation_caption}, image-level captions consistently outperform the simple text baseline across all metrics (+1.2 FPA), demonstrating the advantage of incorporating instance-specific visual descriptions. Nevertheless, the class-level text remains a strong baseline, indicating that even minimal textual supervision is effective.

\noindent \textbf{Caption Cost.} Generating image-level captions takes approximately 2.1 seconds per image on a single A40 GPU. This cost is incurred only once during preprocessing and can be further reduced through parallelization across multiple GPUs.

\begin{table}[h]
\caption{\textbf{Comparison between image-level captions and simple class-level text}. Image-level descriptions improve performance across all metrics, while the simple text baseline remains competitive.}\label{table:ablation_caption}
\centering
\resizebox{0.6\linewidth}{!}{
\begin{tabular}{lccccc}
\toprule
Caption     & FPA ($\uparrow$)  & fine ($\uparrow$) & sub. ($\uparrow$) & basic ($\uparrow$) & TICE ($\downarrow$)\\ \midrule
Simple Text & 62.0 & 64.1    & 84.2   & 93.4  & 20.4 \\
Ours (Image-level) & \textbf{63.2} & \textbf{64.9}    & \textbf{84.5}   & \textbf{93.6}  & \textbf{18.6} \\
 \bottomrule
\end{tabular}
}
\end{table}

\clearpage
\section{Implementation Details}
\label{sec:suppl_implementation_details}
For ViT~\citep{dosovitskiy2020image} models, we use ViT-Small for \methodTextViT and \methodSSL and H-CAST-Small~\citep{hcast2025park} for \methodTextCAST to match parameter sizes. 

For \methodTextViT, we insert fully-connected layers to the class token at the 8th, 10th, and 12th layers for basic, subordinate, and fine-grained supervision. The 12th-layer patch features are projected to match the text embedding dimension via an FC layer. 
For \methodTextCAST, hierarchical supervision is applied to the last three blocks, following~\citep{hcast2025park}. Due to low dimensionality in the final block, we align text features with the features of the second block.
For Text-Attr methods, CLIP-ViT-B/32 is used to extract text embeddings, which remain frozen during training.

In \methodSSL, we apply a shared MLP to the class token from the final (12th) layer, followed by three separate linear classifiers for basic, subordinate, and fine-grained supervision. When combined with Text-Attr, we additionally project the class token through a linear layer and align it with the corresponding text feature.

For hierarchical classification baselines, HRN~\citep{hrn2022chen} and H-CAST~\citep{hcast2025park}, we follow their original training protocols and retrain them under our free-grain setting. We extend HRN to handle missing labels at two levels instead of one. For H-CAST, we provide supervision using the available labels at each corresponding level.
Full hyperparameter configurations are provided in Table~\ref{tab:suppl_implementation_details}.

We train all models for 100 epochs, except for \dataINReal, which are trained for 200 epochs due to the larger scale.
All experiments were conducted on an NVIDIA A40 GPU with 48GB memory. We used a single GPU for all experiments, except for \dataINReal, which was trained using 4 GPUs.

\begin{table}[h]
\vspace{5mm}
\caption{\textbf{Hyperparameters for training \methodTextViT, \methodTextCAST, and \methodSSL.} We follow the training setup of H-CAST~\citep{hcast2025park} for Text-Attr methods (\methodTextViT and \methodTextCAST), and adopt the settings of CHMatch~\citep{wu2023chmatch} for \methodSSL.
}
\label{tab:suppl_implementation_details}
\centering
\small
\begin{tabular}{l|ccc}
\toprule
Parameter              & \methodTextViT        & \methodTextCAST &   \methodSSL             \\ \midrule
batch\_size            & 256             & 256                        & 128\\
crop\_size             & 224             & 224                        & 224\\
learning\_rate         & $5e-4$       & $5{e-4}$                  & $1e-3$ \\
weight\_decay          & 0.05            & 0.05                       & 0.05\\
momentum               & 0.9             & 0.9                        & 0.9\\
warmup\_epochs         & 5               & 5                          & 0 \\
warmup\_learning\_rate & $1e{-6}$       & $1e{-6}$                  & N/A\\
optimizer              & Adam            & Adam                       & SGD\\
learning\_rate\_policy & Cosine decay    & Cosine decay               & Cosine decay\\
$\alpha$ (weight for $\mathcal{L}_{\text{text}}$)               & 1             & 1 & 1 (for +Text-Attr)  \\ 
\bottomrule
\end{tabular}
\end{table}


\clearpage
\section{Full Losses for Taxonomy-guided Semi-Supervised Learning (Taxon-SSL)}\label{supple:sec:taxon-ssl}
In this we provide full details of Taxonomy-guided Semi-Supervised Learning (Taxon-SSL).
As described in Sec.~\ref{method2:semi}, following standard practice~\cite{wu2023chmatch}, our classifier consists of a shared feature extractor $f_{\text{feat}}$ and level-specific heads $\{h_l\}_{l\in \mathcal{S}_x}$. For supervised samples with known labels $y_1, \dots, y_L$ across $L$ taxonomy levels, we apply a per-level hierarchical supervision loss $\mathcal{L}_{\text{sup}}$:
\begin{equation}
    \mathcal{L}_{\text{sup}}= \sum_{l=1}^{L} \mathbbm{1}_{\{y_l \text{ exists}\}} \cdot \mathcal{L}( h_l(f(x)), y_l ).
\end{equation}
For unlabeled levels, we generate a weakly augmented image ($\mathcal{W}(x)$) and a strongly augmented version ($\mathcal{S}(x)$).
Confident predictions from $\mathcal{W}(x)$ at each levels become pseudo-labels for $\mathcal{S}(x)$, denoted as $\mathcal{PL}_l(x)$. Concretely, our pseudo-labeling loss $\mathcal{L}_{\text{pl}}$ is
\begin{equation}
    \mathcal{L}_{\text{pl}}= \sum_{l=1}^{L} \mathbbm{1}_{\{\mathcal{PL}_l(x) \text{ is over confidence threshold}\}} \cdot \mathcal{L}( h_l(f(\mathcal{S}(x))), \mathcal{PL}_l(x) ).
\end{equation}
Confidence thresholds follow the percentile-based schedule introduced in CHMatch~\cite{wu2023chmatch}.

Lastly, to reinforce hierarchical structure in the learned representation, we construct a taxonomy-aligned affinity graph.
 Within each mini-batch $\mathcal{B}$, we construct affinity graphs $W^l$ for each hierarchy, where $W^l_{ij}=1$ if the $i$th and $j$th image have the same pseudo-labels and $W^l_{ij}=0$ otherwise. Then the taxonomy-aligned affinity graph $W$ is defined by
\begin{equation}
    W_{ij} = \left\{
            \begin{aligned}
                &1\quad\text{if } W^1_{ij}=...=W^L_{ij}=1\text{,} \\
                &0\quad\text{otherwise.}
            \end{aligned}
            \right.
\end{equation}
We pull together positive pairs where $W_{ij}=1$ and push apart negative pairs where $W_{ij}=0$. Formally, $\mathcal{L}_{\text{tacl}}$ is defined by
\begin{equation}
    \mathcal{L}_{\text{tacl}}= -\frac{1}{\sum_{j}W_{ij}} \cdot  \text{log}\frac{\sum_{j}W_{ij}\text{ exp}((g(f(x_i))\cdot g(f(x_j))')/t)}{\sum_{j}(1-W_{ij})\text{ exp}((g(f(x_i))\cdot g(f(x_j))')/t)},
\end{equation}
where $i$ is the index of current image, $g$ is the projection head for $\mathcal{L}_{\text{sup}}$, and $t$ is the temperature hyperparameter. 

Our final training objective is: 
\begin{equation}
    \mathcal{L}_{\text{total}} = \mathcal{L}_{\text{hier}} + \beta\cdot\mathcal{L}_{\text{pl}}+\gamma\cdot\mathcal{L}_{\text{tacl}},
\end{equation}
 where $\beta$ and $\gamma$ control the relative contributions of the loss terms, and we set both to 1 in all experiments.

\clearpage 
\section{Related Work}\label{sec:suppl_related_work}
\textbf{Hierarchical recognition} has been used to refer to different tasks.
A large body of work leverages taxonomies only to improve \textbf{leaf-node prediction}~\citep{karthik2021nocost, zhang2022use, zeng2022learning, garg2022mistake}. Thus, evaluation in these works typically relies on top-1 accuracy or mistake severity at the leaf level.
These methods predict a \textit{single fine-grained label} and assume the full hierarchy can be recovered from it.
However, fine-grained prediction is often impossible in real-world scenarios due to visual ambiguity, leading these models to produce uninformative outputs.

In this paper, we focus on \textbf{full-taxonomy prediction}~\citep{chang2021fgn, wang2023consistency, biderection2024jiang, hcast2025park}, where the model must output labels at all levels of the taxonomy.
This setting introduces cross-level inconsistency, as predictions across levels must align with the hierarchical structure.
Recent work~\cite{tan2025hiervqa} shows that such inconsistencies arise even in large vision–language models like GPT-4o~\cite{hurst2024gpt4o} and Qwen2.5-VL-72B~\citep{bai2025qwen2}, underscoring the difficulty of the problem.

Existing full-taxonomy prediction methods enforce consistency constraints and demonstrate strong performance~\citep{chang2021fgn, wang2023consistency, biderection2024jiang, hcast2025park}.
However, they assume that all hierarchical levels are fully annotated for every training sample, which is unrealistic.
In practice, annotation granularity naturally varies due to visual ambiguity or annotator expertise—some images may only receive a coarse label (e.g., \textit{bird}), while others have fine-grained labels (e.g., \textit{bank swallow}).

To address this gap, we study a more realistic setting, free-grain learning, where supervision may appear at any level of the taxonomy and the model must infer the complete label path from partially observed labels.
Existing approaches related to partial hierarchical supervision do not fully capture this setting.
\methodHRN~\citep{hrn2022chen} considers partial labels but only in a limited two-level scenario created by randomly removing fine-grained labels, which does not reflect the structured ambiguity found in real taxonomies.
Kim et al.~\citep{kim2023semanticcorrel} also incorporates mixed-granularity labels, but treats them in a flat manner, overlooking the hierarchical relationships essential to our formulation.
These studies were also restricted to small datasets such as CUB~\cite{ref:data_cub200}, whereas our work establishes a general free-grain formulation and provides a large-scale benchmark to study it systematically.

\vspace{1mm}

\noindent\textbf{Hierarchical recognition datasets}. 
Full-taxonomy prediction requires datasets where each hierarchical level is meaningful to predict.
However, large-scale datasets like ImageNet~\cite{ref:data_imagenet} are not designed for this purpose.
Its WordNet~\cite{fellbaum1998wordnet} taxonomy has uneven depths (e.g., \textit{minivan} appears around the 15th level while \textit{teddy bear} appears around the 7th), and contains many abstract nodes such as \textit{entity}, \textit{object}, or \textit{whole} that are \textbf{not useful prediction targets} (Fig.~\ref{fig:imagenet_taxonomy}).
Such a hierarchy supports only leaf-node prediction with mistake-severity penalties~\cite{garg2022mistake}, where the model still predicts a single leaf label and the hierarchy is used merely to score how far an error is from the ground truth.

Because of this limitation, prior full-taxonomy prediction work has relied on small, clean datasets like CUB~\cite{ref:data_cub200} and Aircraft~\cite{ref:data_air}, where the hierarchy is well-defined but the scale is limited.
iNaturalist~\citep{inat21mini} provides a deeper taxonomy, but its scope is restricted to biological species and does not generalize to broad visual domains.

To enable large-scale and general hierarchical recognition, we introduce ImageNet-3L, which provides three semantically meaningful levels without the abstract superordinate nodes (e.g., \textit{entity}, \textit{object}) present in WordNet.
We center the hierarchy on Rosch’s basic-level categories~\cite{rosch1976basic}, the level at which humans naturally identify objects (e.g., \textit{bird}, \textit{vehicle}), and organize categories downward into subordinate and fine-grained levels.
This produces a clean and meaningful three-level taxonomy that focuses on distinctions worth predicting and is well suited for full-taxonomy recognition.
Using this, we create free-grain variants to study hierarchical prediction under varying label granularity.\vspace{1mm}




\noindent\textbf{Long-tailed recognition} has been extensively studied~\citep{LT_liu_2019_CVPR, LT_balsm_2020,LT_ride2021, LT_Park_2021_ICCV, LT_tian2022vl, LT_Park_2022_CVPR, LT_Ha_2023_BMVC, ahn2023cuda, LT_zhao2024ltgc}, mostly focusing on imbalance at a single fine-grained level (\textit{inter-class} imbalance). 
In contrast, we address both \textit{inter-class imbalance} (across classes) and \textit{intra-hierarchy imbalance} (across semantic levels) in a hierarchical setting, where classes themselves may be balanced but label granularity varies across them.
DeepRTC~\citep{wu2020deeprtc} considers taxonomies, but aims to improve inference reliability via early stopping rather than predicting the full taxonomy.\vspace{1mm}

\noindent{\bf Semi-supervised learning} typically combines labeled and unlabeled data at a single fine-grained level~\citep{meanteacher2017, berthelot2019mixmatch, sohn2020fixmatch}. Recent work incorporates coarse labels~\citep{garg2022hiermatch, wu2023chmatch}, but still targets fine-grained accuracy. In contrast, our setting demands consistent prediction across the full taxonomy with heterogeneous supervision, making existing methods not directly applicable.\vspace{1mm}

\noindent\textbf{Weakly-supervised recognition}
 typically aims to predict fine-grained labels when only coarse labels are available during training~\citep{wsrobinson20a, falcon_2024}. These methods assume fully observed labels at a coarse level and focus on improving predictions at a fine-grained level. In contrast, our setting requires handling multi-granularity labels and inferring the full taxonomy.\vspace{1mm}


\noindent\textbf{Large-language models for recognition.}
Recent approaches~\citep{pratt2023platypus, liu2024democratizing, zheng2024llmprompt, saha2024zero_vlm} leverage vision–language models (VLMs) (e.g., CLIP~\citep{ref:clip_2021}) or large language models (LLMs) (e.g., GPT-4~\citep{achiam2023gpt4}) by generating textual descriptions from label names and feeding them into an LLM to improve flat classification.
Their primary goal is to perform label-driven reasoning without training a new visual model.
In contrast, we do not use labels to expand label descriptions.
Instead, we use VLMs to extract textual cues directly from the image, without referencing labels, so that the image encoder can learn visual attributes shared across hierarchical levels when supervision is incomplete.
At inference, our model is image-only and does not rely on VLMs or LLMs, since our goal is hierarchical prediction rather than label-driven reasoning.
%



\end{document}